\documentclass[11pt,oneside,fleqn,reqno,titlepage]{article}
\usepackage{amsmath}
\usepackage{amssymb}
\usepackage[pdftex]{graphicx}
\usepackage{natbib} \citeindextrue
\usepackage{natbib} 
\usepackage{bbm}
\usepackage{url}
\usepackage{bm}
\usepackage{comment}

\DeclareMathOperator*{\argmax}{arg\,max}

\def \sigmatilde{\tilde \sigma}

\def \xhat{\hat x}

\def \hbar{\bar h}

\def \Cbar{\bar C}

\def \Fbar{\bar F}

\def \Pbar{\bar P}

\def \Vbar{\overline V}

\def \Fcal{{\mathcal F}}

\def \Ical{{\mathcal I}}
\def \Jcal{{\mathcal J}}

\def \Xcal{{\mathcal X}}

\newcommand{\bn}{\begin{eqnarray}}
\newcommand{\en}{\end{eqnarray}}
\newcommand{\bns}{\begin{eqnarray*}}
\newcommand{\ens}{\end{eqnarray*}}

\newcommand{\singlespace}{\addtolength{\baselineskip}{-.25\baselineskip}}

\begin{document}

\title{Optimal Learning for Sequential Decisions in Laboratory Experimentation}

\author{Kristopher Reyes, Warren B Powell\\
\\
{\em Department of Operations Research and Financial Engineering}\\
{\em Princeton University} \\
\\
\\
\\
\\
Prepared for SIAM Review\\
\today
\\
\\
\\
\\}

\maketitle

\begin{abstract}
The process of discovery in the physical, biological and medical sciences can be painstakingly slow.  Most experiments fail, and the time from initiation of research until a new advance reaches commercial production can span 20 years.  This tutorial is aimed to provide experimental scientists with a foundation in the science of making decisions.  Using numerical examples drawn from the experiences of the authors, the article describes the fundamental elements of any experimental learning problem.  It emphasizes the important role of belief models, which include not only the best estimate of relationships provided by prior research, previous experiments and scientific expertise, but also the uncertainty in these relationships.  We introduce the concept of a learning policy, and review the major categories of policies.  We then introduce a policy, known as the knowledge gradient, that maximizes the value of information from each experiment.  We bring out the importance of reducing uncertainty, and illustrate this process for different belief models.
\end{abstract}

\pagenumbering{roman}

\clearpage
\tableofcontents
\clearpage


\newpage

\setcounter{page}{1}


\pagenumbering{arabic}

\section{Introduction}

The laboratory sciences offer a landscape of tens of thousands of research projects pursuing everything from new drugs to new devices to new materials that can change society.  However, the pace of research remains painstakingly slow, often taking decades for breakthroughs to emerge (when they do).  A single experiment can take hours to as long as a month.  Years of experimentation pursuing an idea may not pan out (beyond the continual acquisition of knowledge).  But when the breakthroughs do happen, the results can have real impacts on the lives of millions of people.

It is possible to accelerate this process by addressing a different science outside the physics and chemistry of drugs and materials: we are going to focus specifically on the science of making decisions about which experiment to run and how to run it.  We even tackle the question of assessing the risk of pursuing a line of experiments, which may affect key choices in how the experiments are run, or whether they are run at all.  This tutorial will introduce scientists to the process of how to think about making decisions.

Consider the setting.  We face continuous choices such as the concentration of a solvent, the temperature of the process, material flux and pressure, as well as physical parameters such as the diameter of a tube, thickness of a membrane, and the separation of two plates.  In addition there are discrete choices such as the choice of catalyst, drug cocktails, binding sites and probe designs.  A single experiment might require anywhere from five minutes on a robotic scientist, to several days (and up to a month) of laboratory time.  How do we do this as quickly as possible?

These experiments have to be done in sequence (which is not always the case).  We have to decide on the first experiment with nothing more than our knowledge of physics and chemistry.  As data comes in, we have to continue to decide on subsequent experiments, and hope that we reach our goal within our experimental budget.  To do this effectively, we need a process for guiding the sequencing of experiments, which combines an understanding of what the experimental choices are, what we learn from an experiment, and what we are trying to achieve.  Ultimately, what we need is an experimental {\it policy} that guides this process.  Scientists might think of a policy as a kind of protocol that sets the rules for how we choose the next experiment to run.  Throughout our presentation, a policy is simply a set of rules, or a function, that determines the next experiment to run given what we know now.

There are three classes of decisions we are going to address in this tutorial:
\begin{itemize}
\item[1)] The repetitive decisions that have to be made while tuning a process to achieve the best results.  These include:
    \begin{itemize}
    \item[a)] The setting of continuous parameters such as temperatures, pressures,  and concentrations.
    \item[b)] Discrete choices such as catalysts or substituents (to attach to a base molecule to change its behavior)
    \end{itemize}
\item[2)] Process design decisions, which govern the set of steps involved in an experiment.
\item[3)] The decision of whether to pursue an experimental question, as well as key decisions about the experimental process (the types of machinery, the steps in the process, and the budget), taking into account the risk of achieving a goal.
\end{itemize}

The defining characteristic of our problem class is that experiments are expensive.  At a time when there is considerable focus on ``big data,'' we can best describe our problem domain as ``little data.''  In fact, our first experiment has to be made with no data, but we do have the ability to draw on considerable scientific knowledge that is possessed by the scientist, often supplemented by a vast repository of scientific literature.  It is precisely this reason that we tend to use a Bayesian framework that allows us to capture our prior knowledge.  This setting precludes the use of classic design-of-experiments methods, developed largely in the 1960's and 70's, where experiments are designed in advance with the sole goal of fitting a statistical model to a set of data \cite{Montgomery2000a}.

In this tutorial, we outline the five fundamental elements of a sequential learning problem: 1) The state variable, which encompasses our state of knowledge, or belief, that describes what we know about the chemistry and physics of the problem, as well as the physical state (e.g. how machinery is configured, what materials are available) and other information (e.g. the humidity in the lab). 2) The decisions we have to make, which include decisions such as tuning temperatures, pressures and concentrations, discrete choices such as catalysts and drug combinations, and designing the steps of the experimental process. 3) The information we learn from each experiment (or series of experiments).  4) The updating of the state, focusing primarily on updating the state of knowledge.  5) The design of performance metrics and the objective function used to evaluate policies.

The field that involves choosing which experiment to run can be broadly divided into two communities: the classical design-of-experiments (DoE) where a series of experiments are chosen in advance, and then run in batch or sequentially, and sequential design of experiments, where we use the results of one experiment to guide the choice of the next experiment.  Classical DoE focuses on fitting a statistical model (in particular a linear model) where the statistical accuracy of the model depends only on the choice of experimental design variables, and not the outcome of the experiments themselves.  In our work, our experiments are expensive which limits the number of experiments we can run. In addition, we are trying to maximize (or minimize) some metric (and possibly multiple objectives), where previous experiments can be used to guide the next experiment.

There are different communities that focus on the science of learning in a sequential setting, which can be roughly organized into three styles.  The first comes from the applied probability community which first addressed the problem in the form of the {\it multiarmed bandit problem} which describes the (hopelessly artificial) problem of maximizing the return from playing slot machines (known in the U.S. as ``one-armed bandits'') with unknown winning probabilities.  In 1974, the first computationally tractable solution to this problem was introduced in the form of Gittins indices \cite{GiJo74} (see \cite{gittins2011} for a more thorough treatment of this rich literature).  While Gittins indices are not easy to compute, it introduced the basic idea of an index policy, where we compute an index $\nu_x$ for an experiment using experimental control variables $x$ which captures discrete choices (catalysts, solvents, drug regiments) or discretized versions of continuous parameters (temperatures, pressures, concentrations).  The Gittins index has the general form
\bn
\nu^{Gittins,n}_x = \theta^n_x + \Gamma\left(\frac{\sigma^n_x}{\sigma^W},\gamma\right) \sigma^W,  \label{eq:gittinsindex}
\en
where $\theta^n_x$ is our current estimate of how well we will achieve our metric based on $n$ experiments, $\sigma^W$ is the standard deviation of the experimental noise, $\gamma$ is a discount factor, and $\Gamma(\cdot)$ is a special parameter from Gittins theory that involves moderately difficult computation (which has limited the popularity of the method).  The policy guiding the choice of next experiment is to simply choose the largest $\nu^{Gittins,n}_x$ over all possible experiments $x\in\Xcal$ to determine the $n+1^{st}$ experiment to run.  See \cite{PoRy2012}[Section 6.1] for an accessible introduction to this material.

The second community evolved (and continues to evolve) in computer science using an index policy called upper confidence bounding (UCB) with the form
\bn
\nu^{UCB,n}_x = \theta^n_x + 4 \sigma^W \sqrt{\frac{\log n}{N^n_x}}, \label{eq:ucbbasic}
\en
where $N^n_x$ is the number of times we evaluate an experiment run with control variables $x$ within the first $n$ iterations.  The coefficient $4 \sigma^W$ is typically replaced with a tunable parameter.  UCB policies enjoy nice theoretical properties such as bounds on the number of times that we try the wrong ``arm'' (as this community would refer to an experimental choice), but they are simply not well suited to the setting of expensive experiments (they have attracted far more attention in the cyber community, such as finding the best ad to maximize ad-clicks).  See \cite{Bubeck2012} for a good summary of UCB policies and their analysis.

The third community, which we emphasize in this tutorial, focuses on maximizing the value of information from an experiment.  It is such a natural idea that it has been re-invented in different fields with different names.  Geoscientists looking to maximize the return from drilling a well use an idea known as {\it kriging} which gives a value for drilling at a continuous location $x$ (think of this as the latitude and longitude of a well) (\cite{Huang2006}, and \cite{PoRy2012}[Section 16.2]).  This term is similar to the concept of {\it expected improvement} (EI) which captures the value of information from an experiment $x$ (see \cite{Jones1998} and \cite{PoRy2012}[Chapter 5]).  Both of these metrics were designed with the assumption that experiments involve little or no noise.  This idea has evolved to the {\it knowledge gradient} which is the expected value of a noisy experiment, capturing the uncertainty in our original belief about the problem.  We focus on the knowledge gradient in this tutorial because it is best suited to the setting of expensive experiments.  It is also particularly well suited to a process that involves a partnership between the scientist (serving as the domain expert) and the computer guiding the experiment.  The knowledge gradient also brings out all the different dimensions of a learning problem, most notably the belief model which is central to the process.

We begin our presentation with a list of applications in \ref{sec:applications}, followed by an overview of the elements of a learning problem provided in section \ref{sec:modeling}.

We then go through the five elements of a learning model in greater depth.  Section \ref{sec:statevariable} describes state variables, focusing on belief models which are central to experimental sciences, as well as the issue of physical states (secondary to this tutorial).  Section \ref{sec:decisions} discusses the types of decisions that arise in experimental settings and introduces the notion of a policy.  Then, section \ref{sec:whatwelearn} covers what we learn from an experiment, and the types of uncertainty that we have to deal with.  Next, in section \ref{sec:updatingbeliefs} we describe the process of updating our belief model using the information we learned from our last experiment.  Section \ref{sec:objectivefunction} discusses performance metrics and the objective function we use to evaluate policies.

An important dimension of modeling in the experimental sciences is properly handling uncertainty.  Section \ref{sec:understandinguncertainty} presents some thoughts to help us understand uncertainty,  followed by section \ref{sec:searchingforuncertainty} which addresses the question of identifying different sources of uncertainty.  Section \ref{sec:beliefmodels} then addresses the task of developing belief models which are fundamental to the experimental learning process.

We then address the core challenge in experimental sciences of designing what experiment to be run next, which we determine using a rule or {\it policy}, discussed in section \ref{sec:policies}.  We then describe in some depth the concept of the knowledge gradient in section \ref{sec:knowledgegradient}, which maximizes the value of information from an experiment.  The knowledge gradient has proven to be exceptionally useful in the context of expensive experiments where we have to learn as much as possible using limited budgets.

We close our tutorial with a discussion of the process of getting your recommendations implemented (section \ref{sec:implementingrecommendations}), and assessing the risk of a series of experiments (in section \ref{sec:assessingrisk}).  Section \ref{sec:concludingremarks} offers some concluding remarks.

The style of the tutorial is to introduce mathematics as necessary, with an understanding that our primary audience is scientists with little training in probability and statistics.  We have three goals:
\begin{itemize}
\item[1)] To help scientists understand the elements of a learning problem so they approach the process of scientific experimentation in a scientific way.
\item[2)] To provide some simple guidelines that can be used without any further analytical work.
\item[3)] To introduce scientists to the idea of the value of the information from an experiment, which requires thinking through what is learned and how the information is used.
\end{itemize}

\section{Applications}
\label{sec:applications}
Below is a list of applications with which we have been involved.  These help to highlight diverse settings in terms of decisions, uncertainty, learning and metrics.
\begin{description}
  \item[Diabetic drugs] - Approximately 30 percent of patients do not respond well to the most popular drug for controlling blood sugar, Metformin.  Alternative drugs fall into four major categories: sensitizers, secretagogues, alpha-glucosidase inhibitors, and peptide analogs.  Within these groups are subcategories (there are approximately two dozen drugs overall), and each has its own characteristics in terms of blood sugar reduction and side effects, which are unique to each patient.
  \item[Drug discovery] - We have a base molecule with sites where we can connect different substituents, which may consist of as little as a single atom, or more often segments of molecules that change the behavior of the base molecule.  We use the manufactured molecule (drug) and test its ability to kill cancer cells.  The problem is to find the best combination of substituents that kills the most cancer cells \cite{Negoescu2010a}.  A challenge is that there may be 100,000 combinations (or more), yet we only have a budget to test 40 or 50 molecular combinations.
  \item[RNA accessibility] - We need to design molecular probes that are designed to attach to small sequences of nucleotides; if attachment occurs, the probe fluoresces indicating that the RNA segment is accessible. Creating and testing probes is expensive, so the challenge is guiding the process of deciding which probes should be tested to maximize the total fluorescence (which indicates that we are discovering new RNA segments that are accessible).
  \item[Controlled release profiles] - A water-in-oil-in-water (W/O/W) double emulsion system can be used to achieve controlled release of compounds through the optimization of parameters such as surfactant concentrations, droplet parameters, and oil and water volumes to try to match a target release rate.
  \item[Single-walled nanotubes] - A robotic experimental system (ARES) requires tuning four gas flow rates (for $C_2H_4$, $Ar$, $CO_2$ and $H_2$, water vapor pressure and temperature to create single-walled nanotubes where the goal is to maximize the number of single-walled nanotubes (double-walled nanotubes are a common outcome) with the fewest defects.
  \item[Photoconductivity] - The problem was to maximize photoconductivity from light reflecting on a surface covered with nanoparticles that vary in terms of three discrete shapes, as well as different sizes and densities.  The problem requires first making a decision regarding shape and size, after which a series of experiments can be run at different densities.
  \item[Medical decisions] - A patient is complaining of pain in the knee.  After filling out a complete medical history, a doctor may prescribe rehabilitation, pain medications, or (with increasing frequency) complete knee replacement, which requires additional decisions. We need to identify the best medical decisions for a patient with specific attributes, to achieve the best outcome (discharging a patient with acceptable symptoms) at a cost below a given threshold.
\end{description}


These problems illustrate a number of issues that we are going to need to address.  These include:
\begin{itemize}
  \item Discrete decisions (type of diabetes drug, type of catalyst, sequence of amino acides) and continuous discussions (temperatures, concentrations).
  \item Offline learning (experimentation in a lab where a poor result does not matter) and online learning (evaluating medical decisions, testing diabetes medications) where we have to learn in the field.
  \item Lookup table belief models (the performance of a particular drug or catalyst), where we have to develop a belief about each discrete choice (which could be a discretization of a continuous parameter such as concentration).
  \item Correlated beliefs - Observing one diabetes drug, or one sequence of amino acids, helps us learn about other drugs (or sequences) even if we did not directly try that drug (or sequence).
  \item Hierarchical beliefs - Each diabetes drug is a member of a subgroup that is a member of a larger group, providing a hierarchical structure where drugs in a group are related, while members of a subgroup are even more closely related.
  \item Linear belief models - For the drug discovery problem, instead of developing an estimate for the performance of each molecule (out of 100,000), we use a linear statistical model (known as a QSAR model) that approximates the value of each molecular combination using just a few dozen parameters.
  \item Nonlinear belief models - We can create different types of nonlinear models that describe relationships, such as the diffusion of different concentrations of chemicals or the effect of medical decisions on the likelihood of a successful operation.
\end{itemize}

As we progress through the presentation, we will show how to capture all of these elements in a formal model that will help guide the process of structuring an experimental process by identifying the right questions.

\section{Modeling a learning problem}
\label{sec:modeling}


In this section we highlight the five fundamental elements of any sequential decision problem, although here we are going to focus specifically on learning problems.  We are going to introduce some basic notation which will help refine our discussion, and assist with the occasional equation.  Throughout, we let $n$ index the experiment, where $n=0$ represents the time before any experiments have been run.  We generally assume we have a fixed budget $N$.

The elements of a learning system consist of
\begin{itemize}
\item[1)] The state $S^n$ - This captures what we know after $n$ experiments, including our belief about any uncertain parameters, as well as information that might describe the physical state of our system (we have a machine set up to work with a particular catalyst).  In this presentation, we focus primarily on the state of knowledge about unknown parameters.
\item[2)] The decision $x^n$ - This is the decision we make after running the $nth$ experiment, which means it is the settings we use for the $n+1st$ experiment. The index $n$ here indicates when we make the decision; for example, $x^0$ is our first decision which we make before running any experiments.  $x^n$ includes continuous parameters (temperatures, concentrations) and discrete choices (the shape of the nanoparticle for the conductivity experiment, choice of catalyst or metal organic framework, or the choice of drug regimen). It can also include the steps in the experimental process.  We are going to make decisions using a function called the {\it policy} which we denote by $X^\pi(S^n)$ which translates the information in the state variable $S^n$ (our state of knowledge) to an action (experiment) $x^n$.
\item[3)] The information derived from the $nth$ experiment $W^n$ - The variable $W^n$ represents any measurements or observations derived from running the $nth$ experiment.  This might be the photoconductivity of the surface, the strength of a material, the fluorescence of our probe for the RNA molecule, or the reduction in the blood sugar (as well as side effects) from trying a new medication.  Note that the experiment $x^n$ produces the information $W^{n+1}$.  It is important to acknowledge that when we make the decision $x^n = X^\pi(S^n)$ of what experiment to run next, we are choosing $x^n$ before knowing $W^{n+1}$, which means that $W^{n+1}$ is a random variable when we make the decision of what experiment to run.  Further, $W^{n+1}$ typically depends on both the state $S^n$ and our experimental decision $x^n$.  With this notation, we would write the sequence of states, actions and information as
    \bns
    (S^0, x^0=X^\pi(S^0), W^1, S^1=S^M(S^0,x^0,W^1), x^1=X^\pi(S^1), W^2, \ldots)
    \ens
    If we were to repeat this process from scratch starting with the same initial state $S^0$, we would not observe the same outcomes $W^n$, which means the later states $S^n$ would be different, leading to different decisions.
\item[4)] The transition function $S^M(\cdot)$ - We represent the updating of the state using $S^{n+1} =$ \linebreak $S^M(S^n,x^n,W^{n+1})$.  The function $S^M(\cdot)$ is known by various names in the academic literature, but we refer to it as the system model (hence the notation) or the transition function (our most common term).  In this tutorial, the transition function is primarily used to describe the updating of our belief model, but it would also capture information such as the status of a piece of equipment (that might make one experiment easier than another).
\item[5)] The performance metric and objective function - We let $U(s,x)$ be the utility of an experiment given that our state (of knowledge, and physical state if applicable) is $s$, and we run an experiment with control variables $x$.  Note that our utility function might need to combine the cost and time required to complete an experiment, but most important are the metrics that describe the estimated performance.  For example, imagine that $\theta^n_x$ is our current estimate of the number of cancer cells we might kill with drug $x$ based on what we have observed after $n$ experiments.  The estimate $\theta^n_x$ is part of our state variable $S^n$.  Our utility function might be $U(S^n,X^\pi(S^n)) = \theta^n_{x^n}$ where $x^n = X^\pi(S^n))$.  Below we show how to evaluate a policy.
\end{itemize}

With this basic framework, we are now going to step through these again in more detail to bring out the richness of this problem domain.

\section{The state}
\label{sec:statevariable}
The state variable $S^n$ captures three types of information:
\begin{description}
\item[Physical state $R^n$] - This captures the physical state of our experimental system, measured {\it after} the $nth$ experiment ($R^0$ is the initial state).  This might capture the status of a piece of machinery that is ready to run a particular series of tests (e.g. with a particular catalyst), which requires time to be set up to run a different series (e.g. with a different catalyst).
\item[Information state $I^n$] - This variable might capture information such as the temperature or humidity of a lab which influences the experiment.  Again, this is measured after the $nth$ experiment.  As a general rule, we require a decision to change $R^n$, whereas $I^n$ evolves on its own.
\item[Knowledge (or belief) state $K^n$] - This captures our distribution of belief about unknown parameters after the $nth$ experiment.  $K^n$ captures not just our best estimate of a parameter, but also the distribution of what the parameter {\it might} be.  $K^0$ is known as our {\it prior} distribution of belief, and reflects scientific knowledge before any experiments are run.
\end{description}
There are pure learning problems where $S^n$ consists only of our state of knowledge $K^n$, which is our primary focus in this article.  However, physical (and informational) states are a part of many scientific processes.

To understand the knowledge state, we have to first decide on what we have knowledge about!  Imagine that we are trying to develop a material, drug or device that can be characterized by some quality.  Examples might be
\begin{itemize}
\item Materials - We might want to maximize strength, conductivity, transmissivity, lifetime.
\item Drugs - Goals might be maximizing number of cancer cells killed, ability to reduce blood sugar, or simply the ability to attach to a particular receptor.
\item Devices - We may be testing different anode materials to maximize battery storage, or tuning the parameters of an aerosol can to produce a uniform spray, or we may even be testing different rules for guiding the behavior of a driverless electric vehicle to minimize accidents.
\end{itemize}
The simplest belief model assumes that there is a discrete set of decisions (think of choices of catalysts, materials, or drug regimens) which means that we can write $x\in\Xcal = \{1,2, \ldots, M\}$, where we generally assume that $M$ is not too large (e.g. no larger than 1,000).  A lookup table belief model might represent the performance of what we are trying to create (whether it be a material, drug or device) by $\mu_x$, where $\mu_x$ is uncertain.  For example, we might begin with a prior distribution of belief that $\mu_x$ is normally distributed with mean $\theta^0_x$ and variance $\sigma^{2,0}_x$, in which case we would write $\mu_x \sim N(\theta^0_x, \sigma^{2,0}_x)$.

\begin{figure}[tb]
  \center{\includegraphics[width=4.0in]{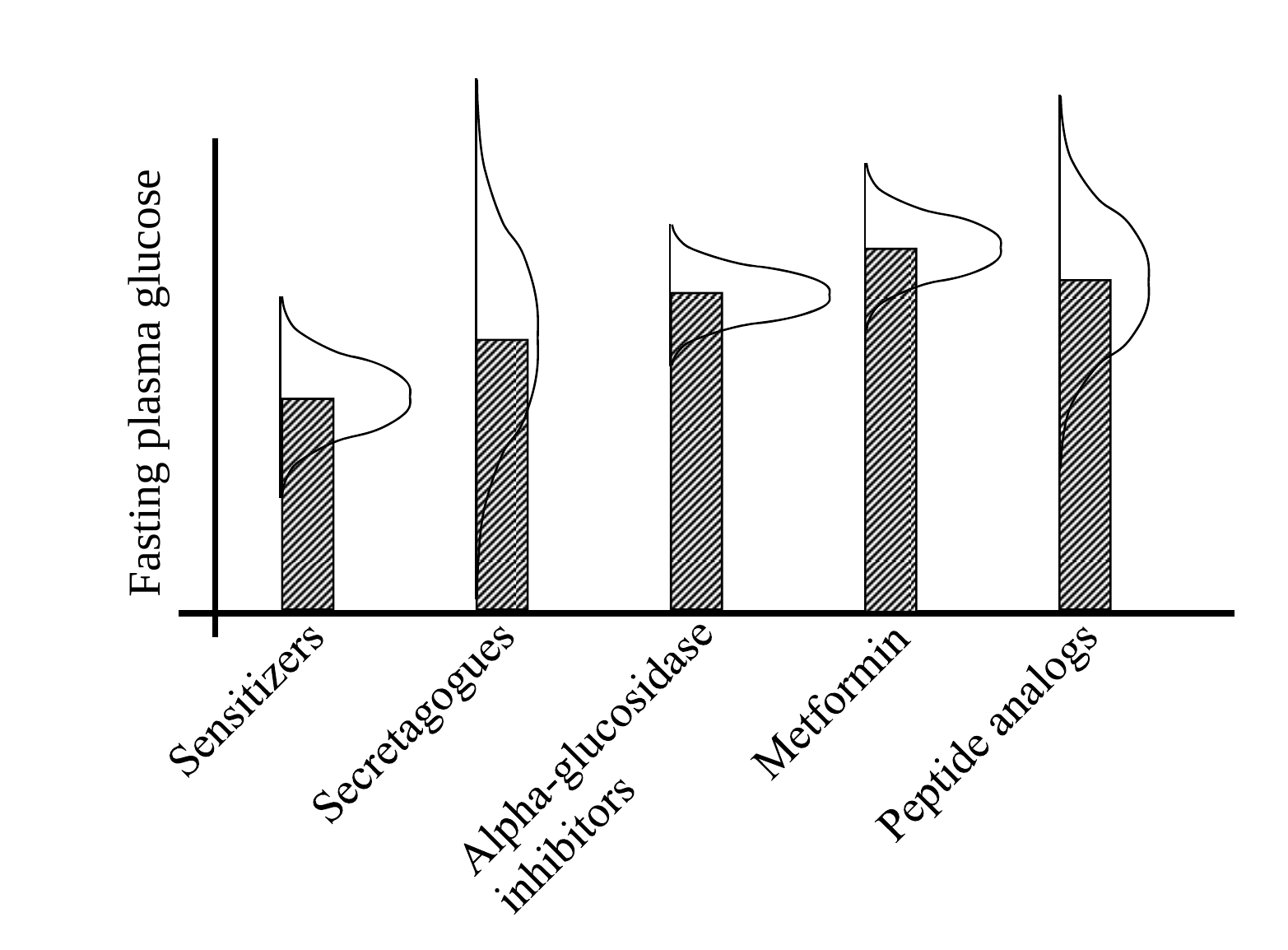}}\\
  \caption{Beliefs about the ability of different medications to lower blood sugar for a particular patient.}\label{fig:basiclearning}
\end{figure}

A lookup table belief model is illustrated in figure \ref{fig:basiclearning} which shows the distribution of belief about different diabetes medications.  Each is characterized by an estimated mean $\theta^0_x$ and standard deviation $\sigma^0_x$ which is used to fit a normal distribution for the range of possible true values for a particular patient.  The challenge faced by doctors is determining what drug to try next, even when a particular drug is working (will another drug work better?).

The idea of treating the truth as a random variable is the defining characteristic of Bayesian statistics: we treat the truth as a random variable, and we start with an initial distribution (in this case the assumption that $\mu_x \sim N(\theta^0_x, \sigma^{2,0}_x)$) called the prior.  The alternative is {\it frequentist statistics}, where estimates are based purely on the results of experiments.  Bayesian statistics is more natural for our setting partly because scientists bring a tremendous amount of domain knowledge to a problem.  In the laboratory sciences, physical experiments are typically too expensive to run the experiments needed to support frequentist estimates.  It is important to exploit domain knowledge to guide the early experimental process before we have collected enough data to build a model purely from observations.

Later we are going to introduce a number of generalizations of this simple belief model (which rarely arises in practice).  Although there is a vast range of statistical models that we can use to represent our belief, in section \ref{sec:beliefmodels} we are going to illustrate lookup tables with correlated beliefs (where our belief about $\mu_x$ is correlated with $\mu_{x'}$), linear belief models (where we use linear regression to approximate the relationship between the performance and the control variables $x$), and nonlinear belief models.

It is particularly important to remember at all times that our belief model is a probability distribution, not a point estimate.  This is key, because the whole point of doing an experiment is reducing our uncertainty in determining the truth.  In section \ref{sec:updatingbeliefs} we show how to update our simple lookup table belief model, and defer to section \ref{sec:beliefmodels} the updating of the more general belief models.

\section{Decisions}
\label{sec:decisions}
We let the variable $x = (x_1, x_2, \ldots, x_K)$ represent the different decisions that govern an experiment ($x_1$ might be the concentration of a solvent, $x_2$ could be temperature, $x_3$ could be the choice of solvent, and so on.  This notation hides a wide range of different types of decisions, which can include
\begin{itemize}
\item Discrete alternatives - Here we assume that $x\in \Xcal = \{1, 2, \ldots, M\}$, where $M$ is not too large.  For example, this could be a decision about different types of catalysts, solvents, metal organic frameworks or the shapes of nanoparticles.  In health, it could be testing medications, dosages, or whether to run a particular test.  It is also possible to discretize what would otherwise be a continuous decision.
\item Multiattribute alternatives - A generalization of discrete alternatives are those where an alternative might be characterized by multiple attributes that makes hierarchical classification possible. For example, sulfonylureas and glinides are two types of diabetes medications that are both types of secretagogues.
\item Continuous controls - $x$ could be one or more continuous decisions such as temperatures, pressures, concentrations, angles, and densities.  $x$ can be a scalar or multidimensional vector.
\end{itemize}

Decisions can also be characterized by other dimensions such as
\begin{itemize}
\item Cost - Costs may vary depending on the material being used, or the facility required to complete an experiment.
\item Time - It may take an hour to test the conductivity of different densities of nanoparticles, but a day or more to try out different shapes which have to be ordered or fabricated.  Similarly, a lab may have the materials on hand to create one RNA probe, but may have to wait two days to receive the materials for another probe.
\item Noise - We may reduce the noise of an experiment by choosing, for example, between atomic force microscopy vs. scanning electron microscopy, or simply repeating the experiment seeral times and averaging.
\item Setup time/cost - A series of experiments with a new material or on a new machine may require setup time (and possibly cost).  This is also known as a switchover cost.
\item Process design - While we primarily focus on decisions that are being made within the context of a fixed sequence of steps, we could choose to use a different set of steps.
\item Sequential vs. batch - There are settings where experiments have to be run in sequence, but there are also many settings where they can be run in batch.
\item Nested decisions - Imagine that you first have to find the size and shape of a nanoparticle, and then run tests on density.  In fact, it might be the case that the density experiments can be run in batch, created a hybrid nested decision process combining sequential and batch decisions (see \cite{wang2015}).
\end{itemize}

While we primarily focus on well-defined control variables $x$ in a well-defined set of possible values $\Xcal$, scientists dealing with experiments that are simply not working have to consider the decisions they are not even thinking about.  Scientists routinely describe breakthroughs due to ``serendipity'' which are little more than accidental ``decisions'' where an experiment was not properly conducted, and yet resulted in a breakthrough.  These are the decisions that have not even been recognized as part of the set of possible choices.

While we would like to pose our problem as one of making the best decisions, the correct way to approach this problem is one of finding the best {\it policy}, which is the rule for making a decision.  An example of a policy is one called ``pure exploitation,'' which means we always choose the design that we think (given what we know) works the best.  To illustrate this, we have to introduce a form of belief model.  The simplest (known as a {\it lookup table} belief model) uses an estimate of the performance of each of a specific set of choices to run an experiment.  Assume that there are $M$ different possible choices of $x$ (in reality, this number may be quite large or even infinite, but we are going to assume that we have identified a not-too-large set of possible experiments we are willing to run).

Now let $\theta^n_x$ be our current estimate (after running $n$ experiments), of the performance when we run our experiment using $x$ (concentrations, temperatures, catalysts) as our set of experimental choices.  Given $\theta^n_x$ for each $x$ in our set of choices $x_1, \ldots, x_M$, a pure exploitation policy simply runs the experiment that seems as if it would produce the best results.  We write this mathematically as
\bn
X^{Eplt}(S^n) = \argmax_{x\in\Xcal} \theta^n_x.  \label{eq:exploitation}
\en
where $\argmax$ means the value of $x$ that corresponds to the largest $\theta^n_x$.  The policy $X^{Eplt}(S^n)$ means simply choosing the design for experiment $n+1$ that looks like the best design given what we know after the $nth$ experiment.

An alternative policy is pure exploration, where we pick the settings of the control variables $x$ completely at random (presumably within some reasonable region).  The problem with pure exploration is that it completely ignores how well something might work.  As a rule, neither pure exploitation nor pure exploration will work well, but striking a balance requires that we understand what

A pure exploitation policy will almost always work poorly, because they result in a tendency to quickly become stuck in a design that only seems to be best (and which may not work at all).  What this tutorial does is to formalize the process of identifying effective experimental policies.


\section{Experimental outcomes}
\label{sec:whatwelearn}
Once we have decided on the experiment we are running, we then run the experiment and observe the results.  These might include
\begin{itemize}
   \item Performance indicators (strength of a material, conductivity, reflectivity, transmissivity, reduction in blood sugar, reduction in cancer cells).
   \item Flaws in the material or product we are trying to produce.
   \item Time required to complete an experiment.
   \item Cost of an experiment.
\end{itemize}

Imagine we are trying to maximize the conductivity of a material.  It is tempting to think that the outcome is the conductivity that we have achieved in our latest experiment.  Actually, the outcome of an experiment is the deviation from what we expected. For example, if we ran our latest experiment with settings $x=x^n$ (concentration of materials, temperature and timing of the heating process), our best estimate of the results of an experiment would be given by $\theta^n_x$.  The actual outcome of the experiment is $W^{n+1}_x$.  What we really learned is the difference $W^{n+1}_x - \theta^n_x$.

Scientists struggle with identifying the causes for variability from one experiment to the next.  Some examples from physical experiments include
\begin{itemize}
  \item Moisture in a chamber
  \item Natural variations in the mixing of materials
  \item Contaminants in a mixture
  \item Variations in the desired temperature
  \item Oxidation in the surface of a catalyst
  \item Variations in the desired concentrations
  \item Shifts in nozzles and sensors
\end{itemize}
Rather than identifying each source of variability, it is common practice to roll the collective contribution of these sources (known and unknown) into a combined experimental noise.  If $\mu_x$ is the (unknown) ``true'' performance of an experiment (think of this as the average if an experiment were repeated a million times) and $\epsilon^{n+1}$ different due to experimental variability, what we observe is
\bns
W^{n+1}_x = \mu_x + \epsilon^{n+1}.
\ens
The experimental noise  $\epsilon^{n+1}$ ``hides'' the true performance $\mu_x$ from running an experiment with control variables $x$.  We begin with an initial belief $\theta^0_x$, and we try to learn $\mu_x$ through repeated experiments, but experimental variations (expressed as the noise $\epsilon^{n+1}$) keeps us from learning $\mu_x$ exactly.

As of this writing, we do not have a strong handle on the modeling of experimental variability. For example, it is fairly standard to view experimental noise as if we were rolling some dice each time we run an experiment.  In practice, experiments can run in streaks, changing behaviors that might be attributed to a particular batch of materials, the setting of a nozzle, or the humidity in the lab.

\section{Updating beliefs}
\label{sec:updatingbeliefs}
For the moment, we are going to just describe the updating process for our basic lookup table model with independent beliefs.  To simplify the algebra a bit, we are going to introduce the idea of the {\it precision} of a distribution, which is simply one over the variance.  Thus, the precision of our experimental noise is given by
\bns
\beta^W = 1/(\sigma^W)^2.
\ens
Similarly, the precision of our belief about $\mu_x$ would be given by
\bns
\beta^n_x = 1/\sigma^{2,n}_x.
\ens
Now assume we decide (after completing the $nth$ experiment) that we are going to run the $n+1st$ experiment using settings $x^n = x$, after which we observe $W^{n+1}$.  Our updated estimates of $\theta^n_x$ and $\beta^n_x$ (for all possible values of $x$) are given by
\bn
\theta^{n+1}_{x} &=& \begin{cases} \frac{\beta^n_x \theta^n_x + \beta^W_x W^{n+1}_x}{\beta^n_x + \beta^W_x} & \mbox{if $x^n = x$}\\
                                  \theta^n_x & \mbox{otherwise,} \end{cases}\label{eq:bayesmeanx} \\
\beta^{n+1}_{x} &=& \begin{cases} \beta^n_x + \beta^W_x  & \mbox{if $x^n = x$}\\
                                  \beta^n_x & \mbox{otherwise.}\end{cases} \label{eq:bayesprecisionx}
\en
We see that the updated value for $\theta^{n+1}_x$, when $x=x^n$ as given by \eqref{eq:bayesmeanx}, is a weighted sum of the prior estimate $\theta^n_x$ and our latest observation $W^{n+1}$ (when we run experiment $x=x^n$).  The updated precision (for $x=x^n$) is simply the sum of the precision of the previous estimate, $\beta^n_x$, and the precision of our experiment $\beta^W$ (we can use $\beta^W_x$ if we think the experimental noise depends on the specific parameter settings for the experiment).  Thus, the precision in our belief about any experiment that we run always gets better, while our beliefs about all other possible experiments remain the same.

We emphasize that this simple model has not applied to any problem we have actually worked on, but it helps to illustrate our transition function for the belief model.

We may have to model a physical state $R^n$, and possibly an information state $I^n$.  For example, if our process is set up to test one type of catalyst (such as cobalt) and our decision is to switch to iron, then we have to incur a setup time and cost.  This change would be represented by our physical state $R^n$, which would capture which catalyst we are currently handling.  If $x^n$ calls for a new catalyst, then we have to model the time and cost, and capture this change in $R^{n+1}$.

\section{Objectives}
\label{sec:objectivefunction}
We now address the problem of deciding how to evaluate how well we are doing.  We need to consider the following dimensions:
\begin{itemize}
\item[1)] Performance metrics - This is where we capture what we are trying to achieve, whether it is maximizing conductivity, strength, blood sugar reduction or cancer cells killed.
\item[2)] Time and cost - We may wish to minimize time and/or cost, or we may simply have limits on each.
\item[3)] Model fitting - While we may want to maximize some performance metric, we may also be fitting a (typically parametric) model.  If we can do a good job fitting our model, then we can use this model to design the best control variables $x$.
\end{itemize}
We begin by first describing the concept of evaluating a policy, which requires having an appreciation of the inherent variability from running a series of experiments.  We then describe different metrics for evaluating performance.

\subsection{Evaluating a policy}
The process of evaluating a policy is probably foreign to people who actually work in a laboratory.  It requires developing an appreciation of using a process of deciding what experiment to run next (which we call a policy) and then repeating this many times (something that can only be simulated on a computer).

To keep our notation compact, we are going to define a function $F(x^{n-1},W^n)$ that tells us how well we did in the $nth$ experiment given the control parameters $x^{n-1}$ and the outcome we observe $W^n$.  Imagine that we have a budget of $N$ experiments, and let $S^N$ be our state of knowledge after we have run all of our experiments.  We can use our state of knowledge $S^N$ to tell us how well control settings $x$ will work.  For example, using our lookup table, $S^N = (\theta^N_x,\beta^N_x)_{x\in\Xcal}$.  This means we can find the value of $x$ that produces the best performance by simply finding the best value of $\theta^N_x$.

Assuming that we use policy $X^\pi(s)$ to learn state $S^N$, we are going to let $X^{\pi,N}$ be the best design based on the state $S^N$.  We write this using
\bn
X^{\pi,N} = \argmax_{x\in\Xcal} \theta^N_x. \label{eq:bestdesign}
\en
In English, equation \eqref{eq:bestdesign} uses the state $S^N$ (which gives us the estimates $\theta^N_x$) to find the best design.

Now here is the tricky part.  If we run policy $X^\pi$ 100 times (think of 100 labs around the country running the same experiment), we will get 100 different estimates of the final state $S^N$.  This means that $S^N=(\theta^N_x)_{x\in\Xcal}$  is random, which means that $X^{\pi,N}$ is also random.  Let $i$ be the $ith$ simulation of policy $X^\pi$, and let $S^N_i = (\theta^N_{xi})_{x\in\Xcal}$ be the estimates we obtain after the $ith$ simulation.  Finally let $X^{\pi,N}_i$ be the best design based on these estimates.

Now assume that we have chosen controls $x$ and we need to evaluate this.  If we run an experiment with control variables $x$, we will observe performance metrics $W$ which are also random.  If we repeat this experiment 100 times, we will get 100 values of $W$.  Let $W_j$ be the value of $W$ we get from the $jth$ testing with controls $x$. We can get an estimate of the value of controls $x$ by averaging over these experiments, giving us
\bn
\Fbar^{avg}(x) = \frac{1}{J}\sum_{j=1}^J F(x,W_j). \label{eq:averageW}
\en
Next, we need to get the value of the policy.  We do this by averaging over the different solutions we get by following policy $X^\pi$, which we do by computing
\bn
\Fbar^\pi = \frac{1}{I} \sum_{i=1}^I \Fbar^{avg}(X^{\pi,N}_i). \label{eq:averagepi}
\en
Thus, $\Fbar^\pi$ is the average value produced by following policy $X^\pi$, where we use \eqref{eq:averageW} to compute an estimate of the average of how well we would do with a specific design.

This discussion introduces the idea of evaluating an experimental {\it policy} over repeated simulations.  The idea is that if a policy works well (on average, based on \eqref{eq:averagepi}) in a simulated environment, it should work well in a laboratory.  However, there are other ways of evaluating a policy, a question we address next.

\subsection{Performance measures}
The discussion above makes an implicit assumption that a policy should be based on its average performance, evaluated based only on the final design.  It is important to recognize the following choices when evaluating a policy:
\begin{description}
\item[Cumulative vs. final performance] - We need to decide whether we are only interested in the best possible design when we are finished our experimental campaign, or if we would like to maximize performance during the experimental process.
\item[Expectation vs. risk] - Imagine that we can test our policy 100 times (something that we can only do in the simulated world of the computer).  Do we want the policy that does best on average?  Or are we more interested in how often we do not achieve a particular objective?  The first means we are interested in the expected value of a policy, while the second means we are focusing on risk.
\item[Single vs. multiple objectives] - It would be nice if we could evaluate a policy based on a single performance metric, but in practice scientists have to manage multiple objectives.
\end{description}

\subsubsection{Cumulative vs. final performance}
The objective $\Fbar^\pi$ evaluates a policy assuming we are only interested in the final performance, and it evaluates the policy on average, rather than looking at the worst that it may do.  There are good arguments why we may not want to do either of these.

In a laboratory environment, it is common to assume (in the modeling community, that is) that we do not care how poorly we perform in the lab as long as we get the best possible design in the end.  However, while the goal of experimentation is learning, we can make the case that we would like to see some successes in the lab as well.  If we are able to ignore failures in the lab to achieve better performance at the end, then we are interested in ``final performance'' which is often called ``offline learning'' (since we do not have to live with the results of a poor experiments).  If we would like to enjoy successes in the lab, we may choose to optimize ``cumulative performance'' which is sometimes called ``online learning,'' where we try to do the best we can with each experiment.

We write the cumulative performance objective using
\bn
\Fbar^{\pi,cum} = \frac{1}{I}\sum_{i=1}^I\sum_{n=1}^N F(X^\pi(S^{n-1}_i),W^n_i).  \label{eq:cumulativereward}
\en
Here, we accumulate the rewards over all $N$ experiments.  We then repeat this $i=1, \ldots, I$ times and take an average.

\subsubsection{Risk}
Risk is a critical issue in scientific experimentation.  While it seems intuitively reasonable to want a policy that works well on average, we only get to run a series of experiments once.  It is typically the case that the experiments are being run to achieve a metric that exceeds some threshold.  Imagine that we consider the experiment a success if we achieve $\Fbar^{avg}(X^{\pi,N}) \geq F^{thresh}$.  We use the indicator variable $\mathbbm{1}_{\{E\}}=1$ if event $E$ is true.  We use this to count how many times we exceed our threshold, which we write using
\bns
\mathbbm{1}_{\{\Fbar^{avg}(X^{\pi,N}) \geq F^{thresh}\}} = \left\{\begin{tabular}{cl} 1 & \mbox{If $\Fbar^{avg}(X^{\pi,N}) \geq F^{thresh}$} \\
                                                                                0 & \mbox{Otherwise}
                                                                                \end{tabular} \right.
\ens
to count how often we meet our threshold.  Assume for the moment that we focus just on the final design (our terminal reward criterion).  Now we can estimate the probability that we achieve our goal, which we can compute using
\bn
\Pbar^\pi = \frac{1}{I}\sum_{i=1}^I \mathbbm{1}_{\{\Fbar(X^{\pi,N}) \geq F^{thresh}\}}. \label{eq:threshold}
\en
Despite the importance of risk, the literature on design of experiments (batch or sequential) has largely ignored the issue of risk.

\subsubsection{Multiobjectives}
An important issue in the experimental sciences is that scientists may be interested in more than one objective.  For example, a high value of $\Fbar^\pi$ (final or cumulative) may have to be balanced against what is required to achieve this metric.  For example, we may require the use of higher resolution microscopy or more careful analysis to reduce experimental noise. Different objectives can be combined into a single utility function with weights that capture their importance, or they can be simply displayed (graphically if there are only two dimensions, or in a table) so that a scientist can apply subjective judgment to choose which is best.

Multiple objectives can be handled in different ways.  One is to create a utility function that weights different metrics.  For example, we may wish to have a high probability of single-walled nanotubes, but we also want a low presence of imperfections.  We can express a tradeoff by weighting each and adding them together into a common utility function.

A second approach is to focus on maximizing (or minimizing) some metric, subject to a series of targets or thresholds for other metrics.  For example, we might wish to maximize the strength of a material, but wish to run the experiment with a temperature under some threshold.

\subsection{Remarks}
We now have three metrics for evaluating a policy $X^\pi$: the average final reward $\Fbar^\pi$, the average cumulative reward $\Fbar^{\pi,cum}$, and the probability that we reach our threshold $\Pbar^\pi$.  We can use these to test different policies.  For example, earlier we introduced a pure exploitation policy $X^{Eplt}(s)$ (equation \eqref{eq:exploitation}) where we always try what we think is best.  We might want to try a pure exploration policy, call it $X^{Expl}(s)$, which simply picks a design $x\in\Xcal$ at random with probability $1/|\Xcal|$ (this would not work well with a lookup table belief model, but can work reasonably with a parametric model).  If we pick the final reward objective, we just have to compare $\Fbar^{Eplt}$ to $\Fbar^{Expl}$.  Since we have to deal with uncertainty in our simulations, we need to be careful to construct statistical confidence intervals to see if the differences are statistically significant.

At this point, we have established a formal conceptual framework for capturing the different elements of an experimental learning problem.  We then illustrate how to use this framework to compare different experimental policies.  At the heart of any learning problem is the belief model which captures what we know, and how well we know it, which we want to exploit in the design of an effective policy.  So far, however, we have illustrated these ideas in the context of a very simple belief model (lookup table with independent beliefs) which is unlikely to work in real applications.


\section{Searching for uncertainty}
\label{sec:searchingforuncertainty}
One of the most important insights in running experiments is the search for uncertainty.  In a nutshell, there is no point in running an experiment where you are sure of the outcome (although perhaps you were not really sure).

Figure \ref{fig:KGequal} illustrated the tradeoff between what we expect from an experiment and the uncertainty in this estimate for a lookup table belief model.  Figure \ref{fig:twodimensionalkg} shows a different perspective of the knowledge gradient in the setting where the experimental controls $x$ consist of a (discretized) two-dimensional continuous surface, where beliefs about $\mu_x$ and $\mu_{x'}$ are correlated using the declining exponential covariance function given by \eqref{eq:exponentialcovariance}.  The figures on the left show our belief about the function itself, while the ones on the right show the knowledge gradient for each $x$.  As we measure one point, we update our belief (shown on the left), but the knowledge gradient tends to drop in that region (this is usually, but not always the case - the knowledge gradient can increase if our belief increases significantly).
\begin{figure}[tb]
  \center{\includegraphics[width=5.5in]{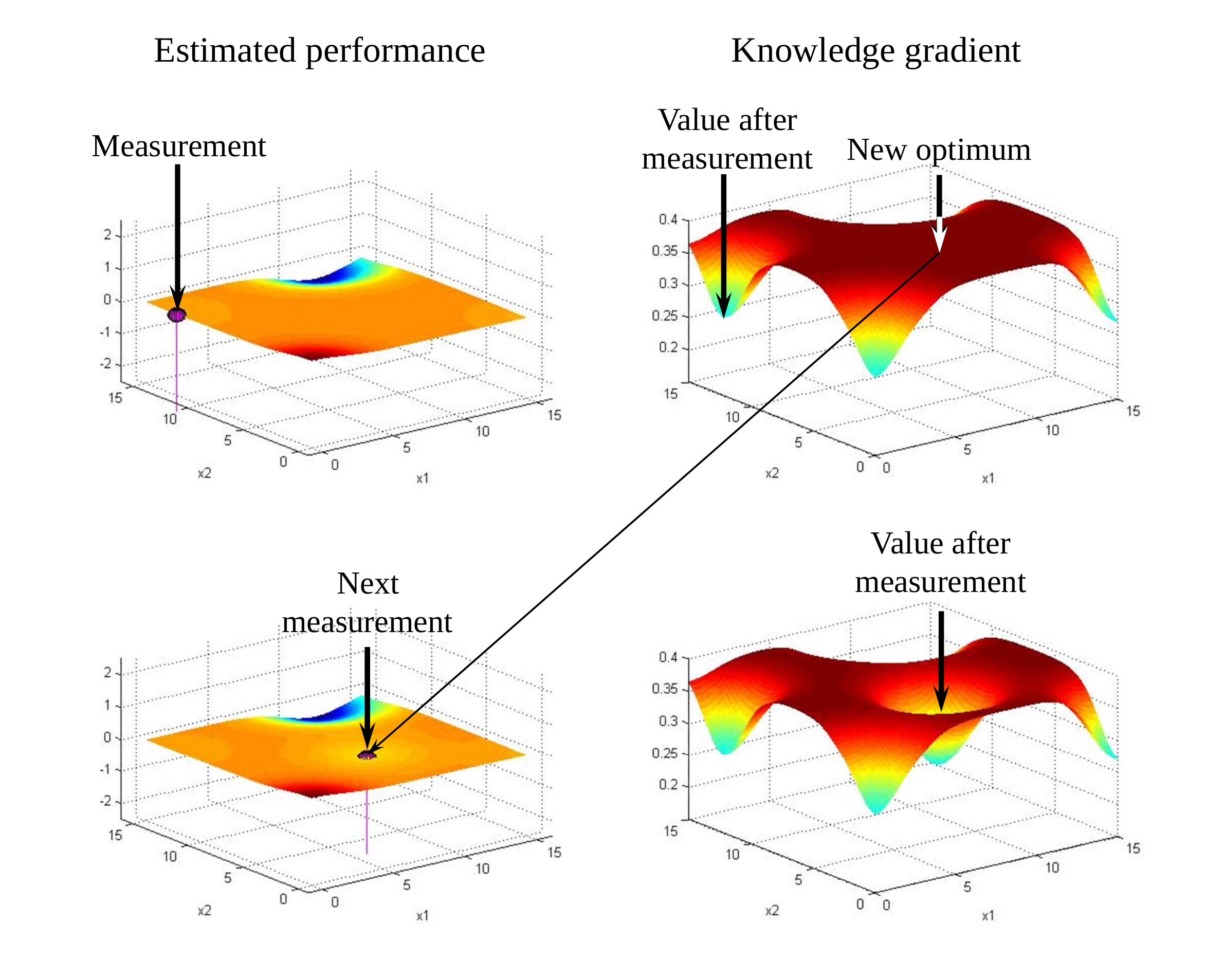}}\\
  \caption{Illustration of prior and KG surface for two-dimensional control vector $x$ with correlated beliefs, demonstrating local learning from an experiment.}\label{fig:twodimensionalkg}
\end{figure}

In experimental settings, uncertainty can exhibit itself in a variety of different ways, as depicted in figure \ref{fig:uncertain}.  Figure \ref{fig:uncertain}(a) shows the simple situation of a series of lines, where there is the greatest uncertainty away from the center of the graph - these are the regions where we would learn the most (recognizing that these may also be the more difficult experiments, requiring the judgment of an experimentalist).  Figure \ref{fig:uncertain}(b) depicts a series of logistics curves, where we are unsure about whether the parameter in question (this might be a temperature or concentration) has a positive or negative impact on the probability of success.  At the same time, we are uncertain about the slope and where the transition begins.  This type of uncertainty suggests starting with some extreme experiments to identify the sign, and then transition to experiments more in the middle to learn about slopes and shifts.
\begin{figure}
\begin{center}
  \begin{tabular}{cc}
  \includegraphics[width=3.0in]{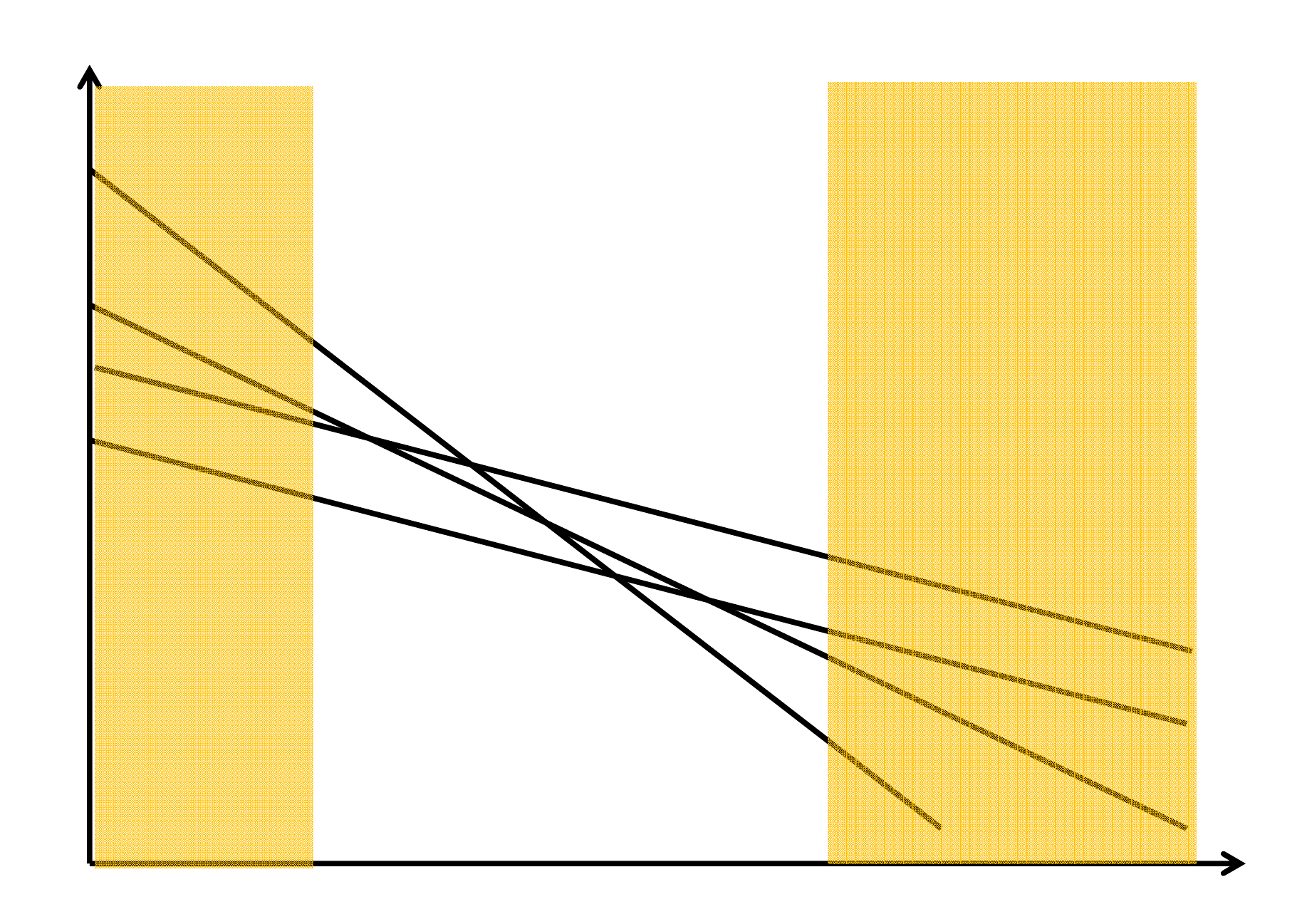} & \includegraphics[width=3.0in]{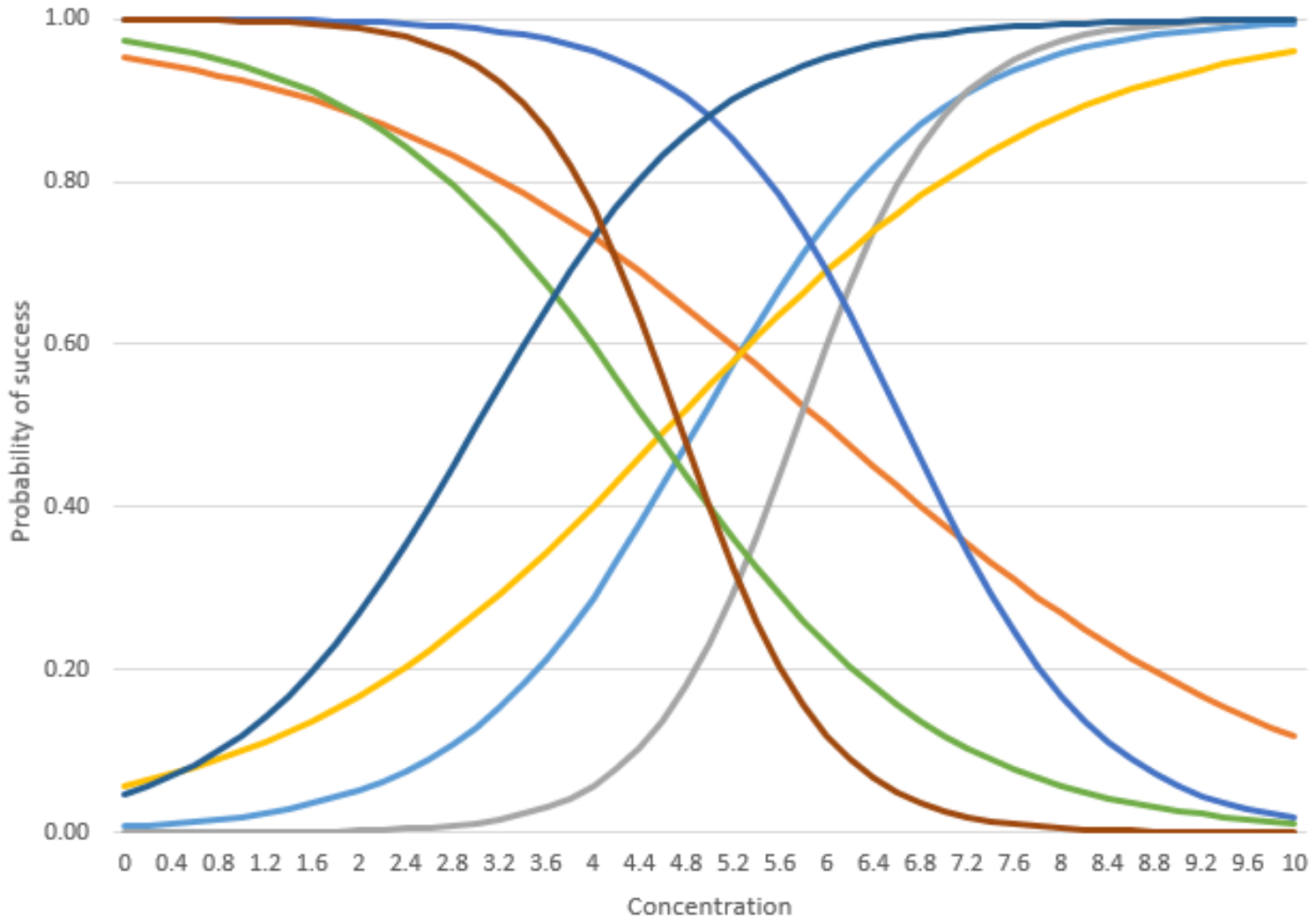} \\
                            (a)                      &                           (b)     \\
  \includegraphics[width=3.0in]{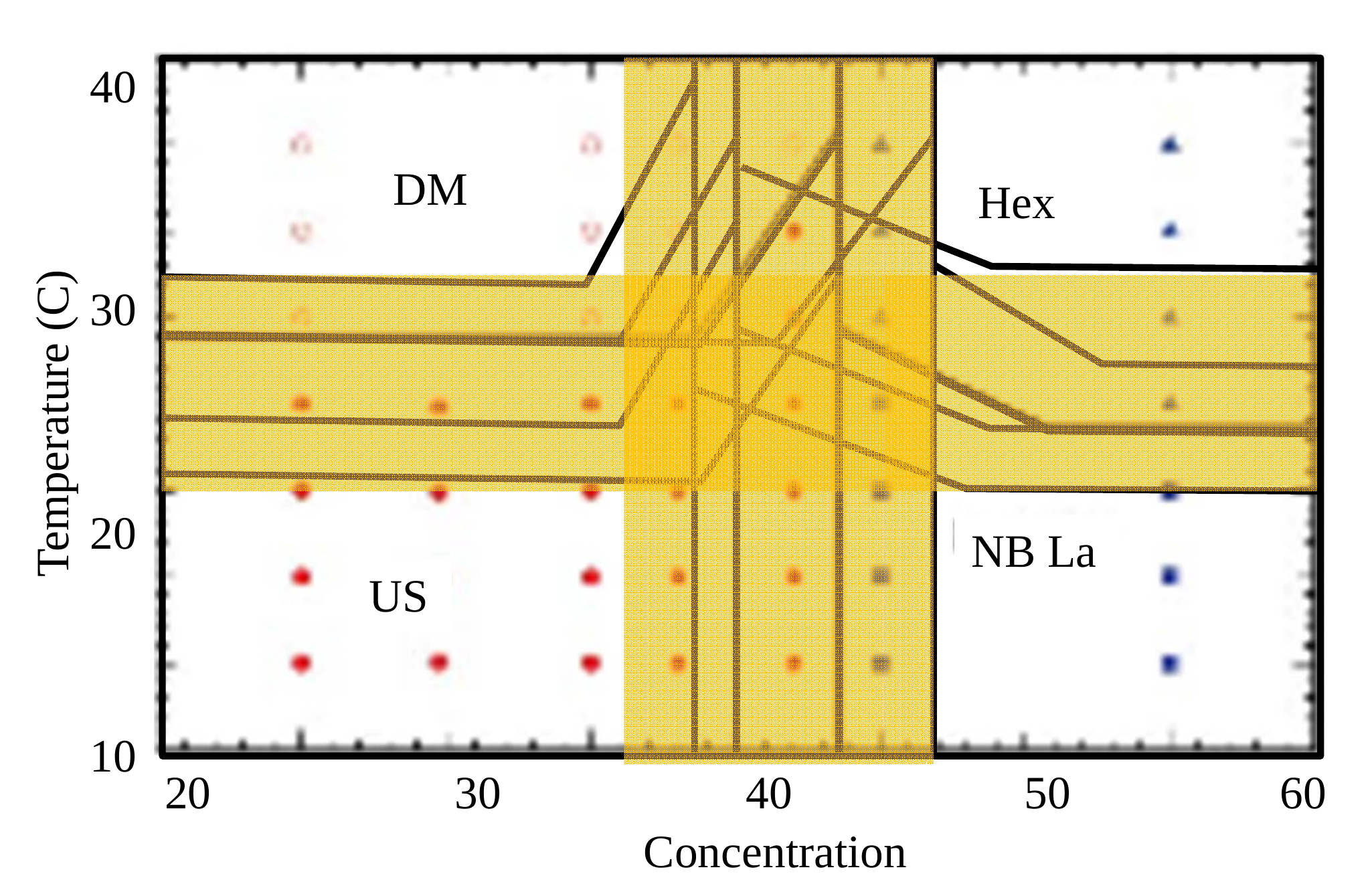} & \includegraphics[width=3.0in]{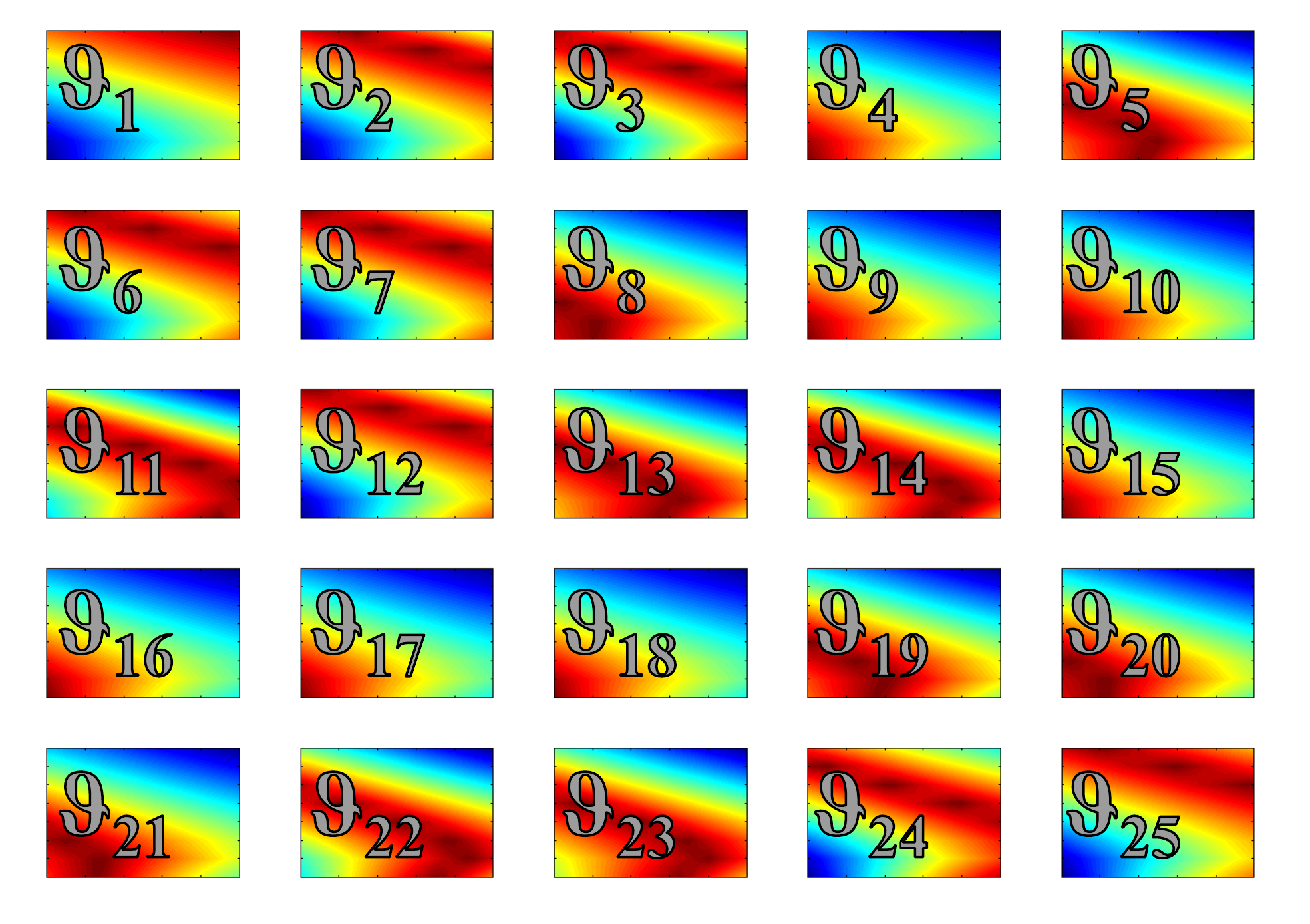} \\
                            (c)                      &                           (d)
  \end{tabular}
\end{center}
\caption{Different flavors of uncertainty:  (a) uncertain lines, (b) uncertain logistic curves, (c) uncertain phase diagrams, (d) uncertain heatmaps.} \label{fig:uncertain}
\end{figure}

Figure \ref{fig:uncertain}(c) depicts a situation where the combination of temperature and concentration can produce four material phases (in this example, each experiment required 1-2 days in a special laboratory).  The lines show where a scientist described his uncertainty about the regions of the phase diagram, which provides a clear indication of where future experiments should be run (where there is the most uncertainty).  Finally, \ref{fig:uncertain}(d) shows 25 heat maps that represent 25 different settings of a set of kinetic parameters.  To help determine which experiment should be run next, we need to find the regions with the greatest variability between the beliefs.

\section{Understanding uncertainty}
\label{sec:understandinguncertainty}
There are a number of different forms of uncertainty when doing experimental research.  As of this writing the ones we have identified from our interactions include:
\begin{itemize}
\item Observational errors - This arises from uncertainty in observing or measuring the state of the system.  Observational errors arise when we have unknown state variables that cannot be observed directly (and accurately), as often happens in the experimental sciences.
\item Experimental noise - Experimental uncertainty, which is distinct from observational uncertainty, refers to variability introduced in the process of running an experiment.  For example, we may be able to perfectly observe whether or not a nanotube is single- or double-walled, but repeated experiments with the same settings produces different results.
\item Control uncertainty - This is where we choose an experimental design or control $x^n$, but what happens is we implement $\xhat^n$ instead of $x^n$.  For example, the experiment might be run at a different temperature or concentration (possibly due to a miscalculation), or the grad student made an error when ordering a compound (a real example of this produced a breakthrough).
\item Inferential (or diagnostic) uncertainty - Inferential uncertainty arises when we use observations to draw inferences about a set of parameters.  It arises from our lack of understanding of the precise properties or behavior of a system, which introduces errors in our ability to estimate parameters, even from perfect measurements.
\item Systematic uncertainty - This covers what might be called ``state of the world'' uncertainty.  The medical community often refers to this as {\it epistemic uncertainty}.   Systematic uncertainty can reflect a missing variable, a bias due to a mechanical problem in an experimental process, or a shift in a disease pattern due to a mutation.
\item Model uncertainty - This covers both uncertainty in the structure of the model we are using, and uncertainty in the parameters of the model (which we capture in our prior).  We note that the model may be locally accurate, implying that there is increasing uncertainty as we move away from a particular region.
\item Goal uncertainty - Uncertainty in the desired goal of a solution, as might arise when a single model has to produce results acceptable to different people or users.  
\end{itemize}

At this point it is useful to mention some of the different distributions that can arise to describe uncertainty:
\begin{itemize}
   \item[-] Gaussian (or normal) distribution - This is the default distribution when modeling continuous errors.  It is typically modeled as having mean zero (if the mean is not zero, then there is a known bias that we can correct).
   \item[-] Exponential or gamma distribution - These are used when the uncertainty has to be kept positive, as might happen when the parameter is positive but the uncertainty in the measurement is large relative to the mean.
   \item[-] Beta distribution - Describes random variables that fall between 0 and 1, which is useful when modeling the probability of a successful outcome.
   \item[-] Bernoulli - Describes random variables that can only be equal to 0 or 1.  This is used when we have uncertainty in an outcome that might be described as success or failure, such as a single-walled nanotube (success) or double-walled nanotube (failure).
   \item[-] Interval - When the range of a parameter is solicited from a domain expert, it is often expressed as an interval.  This might be interpreted as, say, a 95 percent confidence interval from a normal distribution, or as a simple uniform distribution.
   \item[-] Discrete or sampled distribution - We might represent uncertainty in a set of parameters as a discrete set of possible values (as we do below).
   \item[-] Chi-squared - arises when trying to minimize the squared deviation from a target.
\end{itemize}

\section{Searching for uncertainty}
\label{sec:searchingforuncertainty}
One of the most important insights in running experiments is the search for uncertainty.  In a nutshell, there is no point in running an experiment where you are sure of the outcome (although perhaps you were not really sure).

Figure \ref{fig:KGequal} illustrated the tradeoff between what we expect from an experiment and the uncertainty in this estimate for a lookup table belief model.  Figure \ref{fig:twodimensionalkg} shows a different perspective of the knowledge gradient in the setting where the experimental controls $x$ consist of a (discretized) two-dimensional continuous surface, where beliefs about $\mu_x$ and $\mu_{x'}$ are correlated using the declining exponential covariance function given by \eqref{eq:exponentialcovariance}.  The figures on the left show our belief about the function itself, while the ones on the right show the knowledge gradient for each $x$.  As we measure one point, we update our belief (shown on the left), but the knowledge gradient tends to drop in that region (this is usually, but not always the case - the knowledge gradient can increase if our belief increases significantly).
\begin{figure}[tb]
  \center{\includegraphics[width=5.5in]{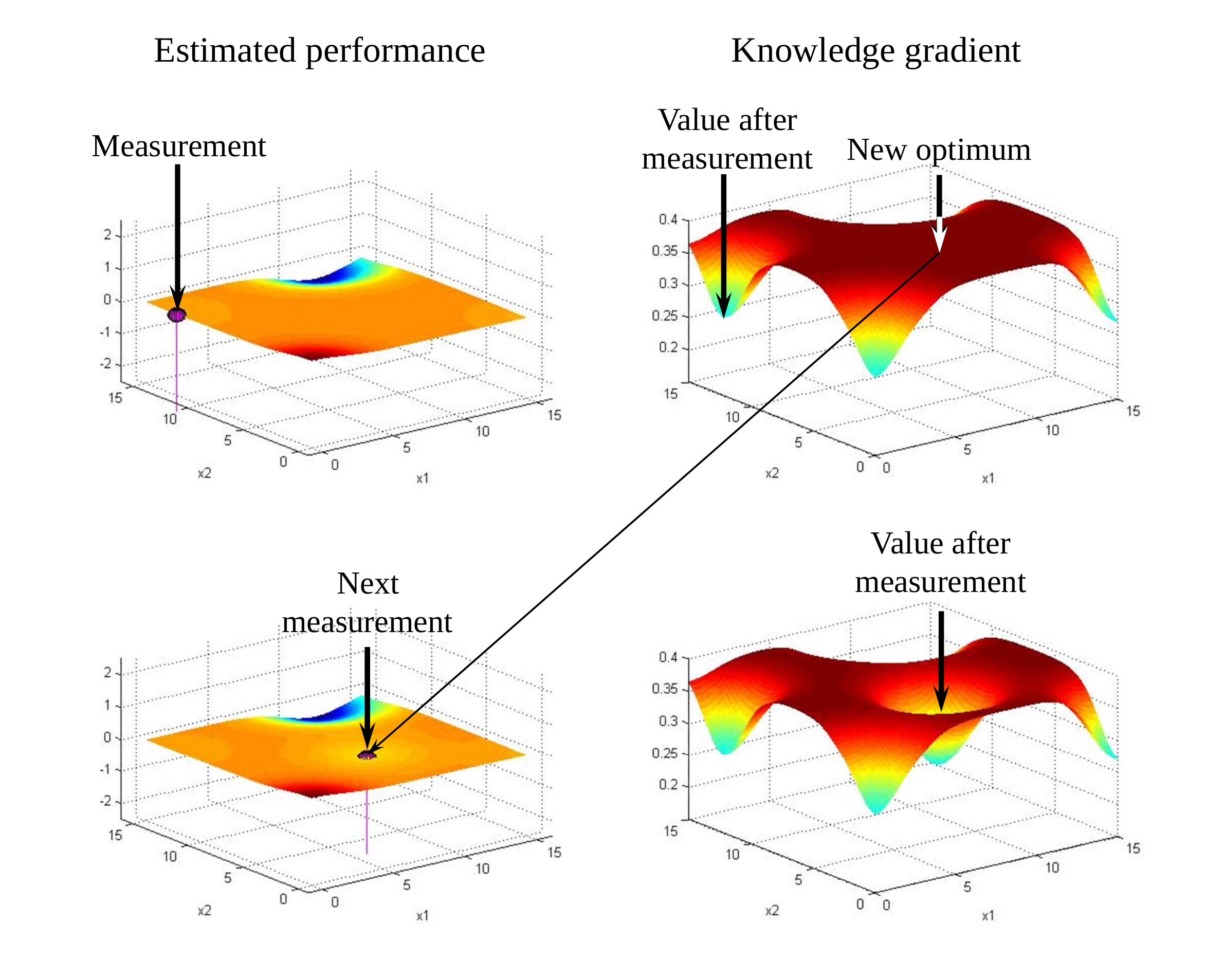}}\\
  \caption{Illustration of prior and KG surface for two-dimensional control vector $x$ with correlated beliefs, demonstrating local learning from an experiment.}\label{fig:twodimensionalkg}
\end{figure}

In experimental settings, uncertainty can exhibit itself in a variety of different ways, as depicted in figure \ref{fig:uncertain}.  Figure \ref{fig:uncertain}(a) shows the simple situation of a series of lines, where there is the greatest uncertainty away from the center of the graph - these are the regions where we would learn the most (recognizing that these may also be the more difficult experiments, requiring the judgment of an experimentalist).  Figure \ref{fig:uncertain}(b) depicts a series of logistics curves, where we are unsure about whether the parameter in question (this might be a temperature or concentration) has a positive or negative impact on the probability of success.  At the same time, we are uncertain about the slope and where the transition begins.  This type of uncertainty suggests starting with some extreme experiments to identify the sign, and then transition to experiments more in the middle to learn about slopes and shifts.
\begin{figure}
\begin{center}
  \begin{tabular}{cc}
  \includegraphics[width=3.0in]{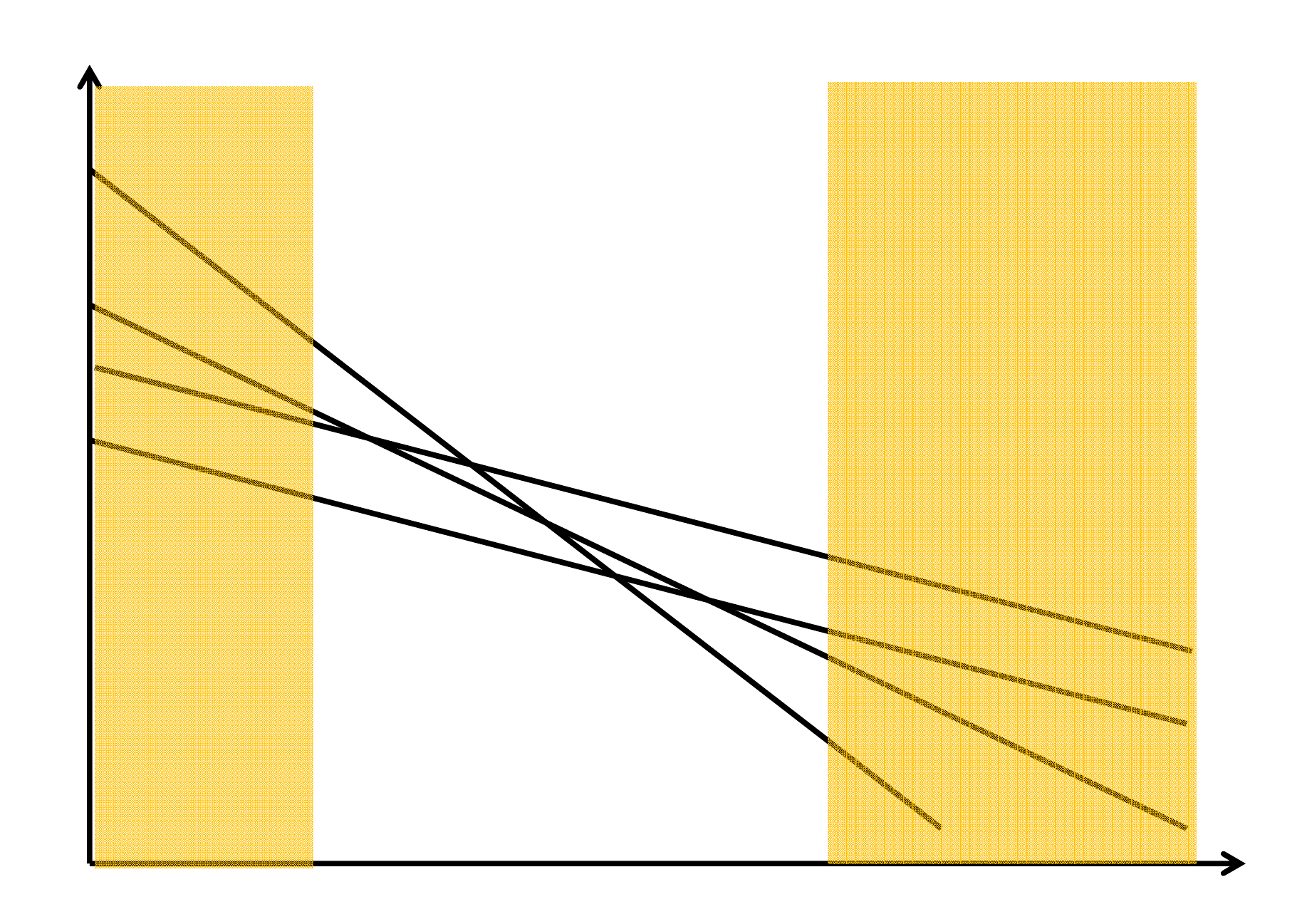} & \includegraphics[width=3.0in]{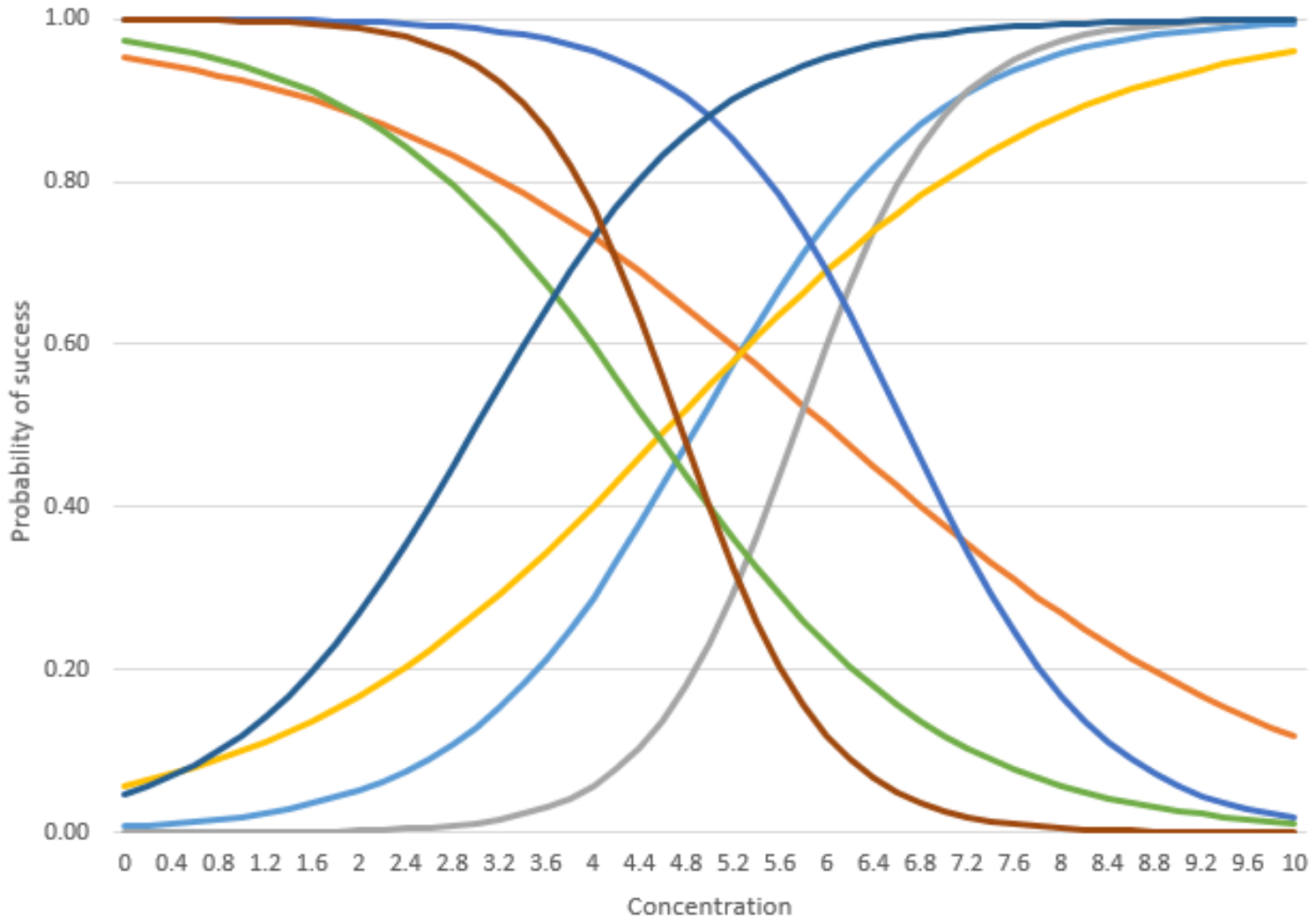} \\
                            (a)                      &                           (b)     \\
  \includegraphics[width=3.0in]{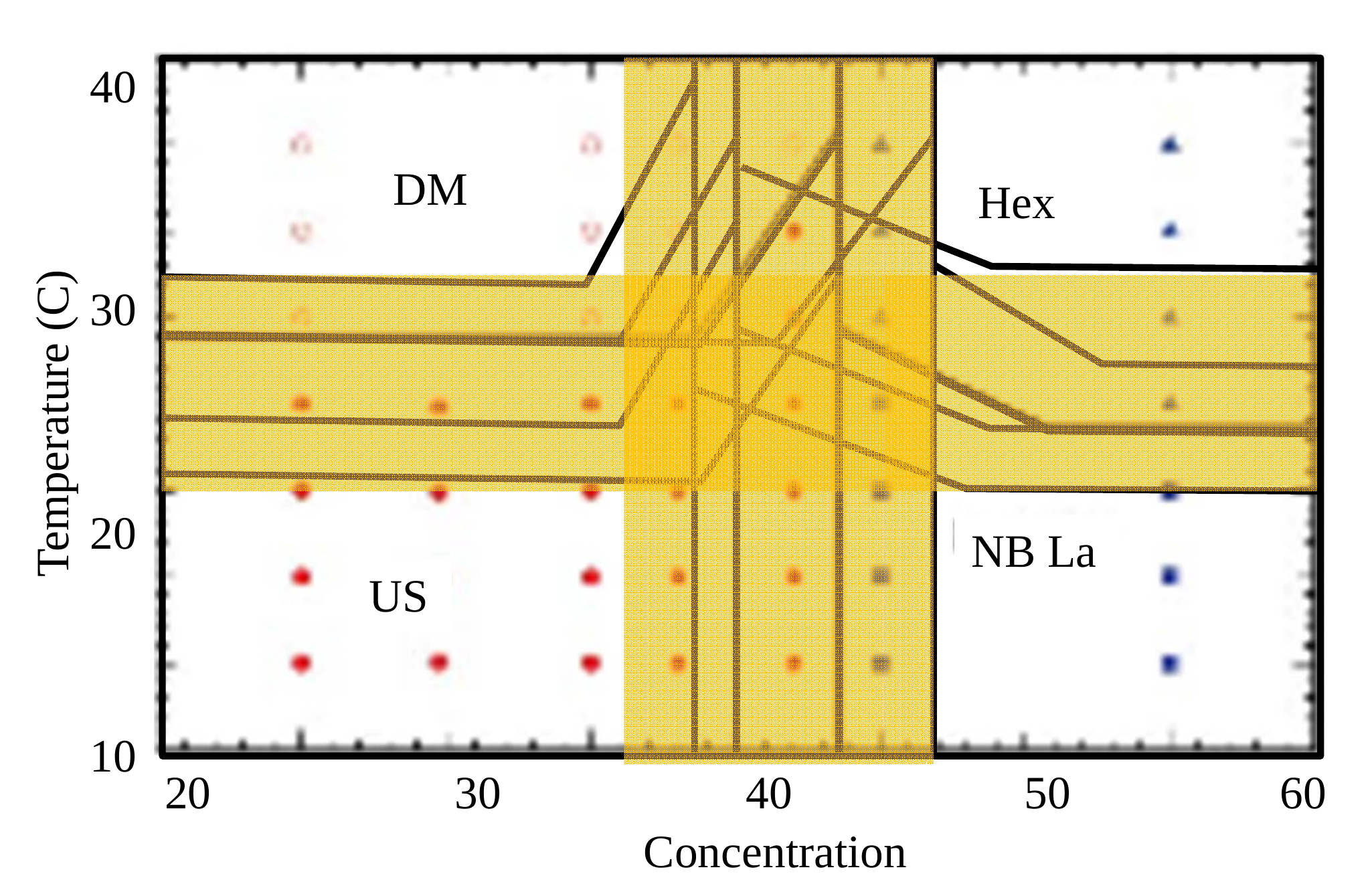} & \includegraphics[width=3.0in]{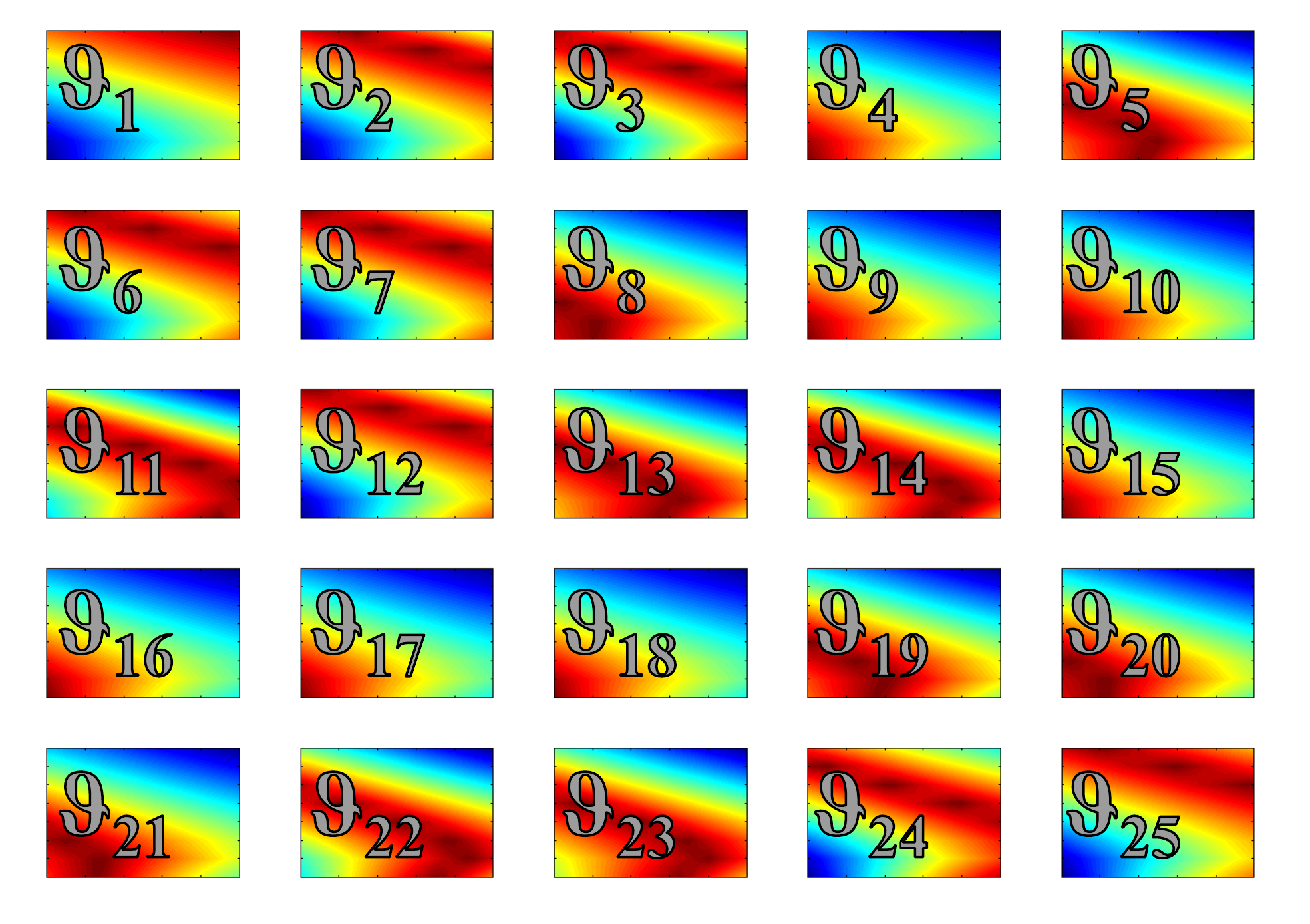} \\
                            (c)                      &                           (d)
  \end{tabular}
\end{center}
\caption{Different flavors of uncertainty:  (a) uncertain lines, (b) uncertain logistic curves, (c) uncertain phase diagrams, (d) uncertain heatmaps.} \label{fig:uncertain}
\end{figure}

Figure \ref{fig:uncertain}(c) depicts a situation where the combination of temperature and concentration can produce four material phases (in this example, each experiment required 1-2 days in a special laboratory).  The lines show where a scientist described his uncertainty about the regions of the phase diagram, which provides a clear indication of where future experiments should be run (where there is the most uncertainty).  Finally, \ref{fig:uncertain}(d) shows 25 heat maps that represent 25 different settings of a set of kinetic parameters.  To help determine which experiment should be run next, we need to find the regions with the greatest variability between the beliefs.

\section{Creating belief models}
\label{sec:beliefmodels}
Arguably the most important dimension of a learning process is the belief model, since this captures what we know, and how well we know it.  The value of an experiment is captured in the belief model, and this is how we make decisions, either about the next experiment, or ultimately in the final design of whatever it is we are trying to make.  A central characteristic of a belief model is that they have to capture the distribution of beliefs so that the uncertainty in the belief is properly captured.  If there is no uncertainty, then there is no need to do further experimentation.

Earlier, we have illustrated our modeling framework using the simplest belief model: a lookup table belief model with normally distributed, independent beliefs.  This means we assume that we have a discrete set of designs $x\in\Xcal = \{1, 2, \ldots, \}$, with a different estimate $\theta^n_x$ for each $x$.  The problem with this approach is that $x$ might be one of a thousand materials, tens of thousands of molecular combinations, or millions of combinations of (discretized) continuous parameters.  With experimental budgets often on the order of several dozen, choosing the best design by creating an estimate $\theta^n_x$ for each $x$ is hopelessly impractical.

In this section, we introduce the three most powerful belief models that we have used in our work: lookup tables with correlated beliefs, linear (parametric) models, and nonlinear models.  We model uncertainty using normally distributed parameters for the first two, and a sampled belief model for nonlinear models (and sometimes even linear, parametric models).  After presenting each belief model, we then give the equations for updating them after each experiment (these equations can be skipped without affecting the understanding of the rest of the tutorial).  After this, we address the important problem of creating the initial belief (known as the prior) before we start any experiments.

\subsection{Lookup tables with correlated beliefs}
\begin{figure}[tb]
  \center{\includegraphics[width=5.0in]{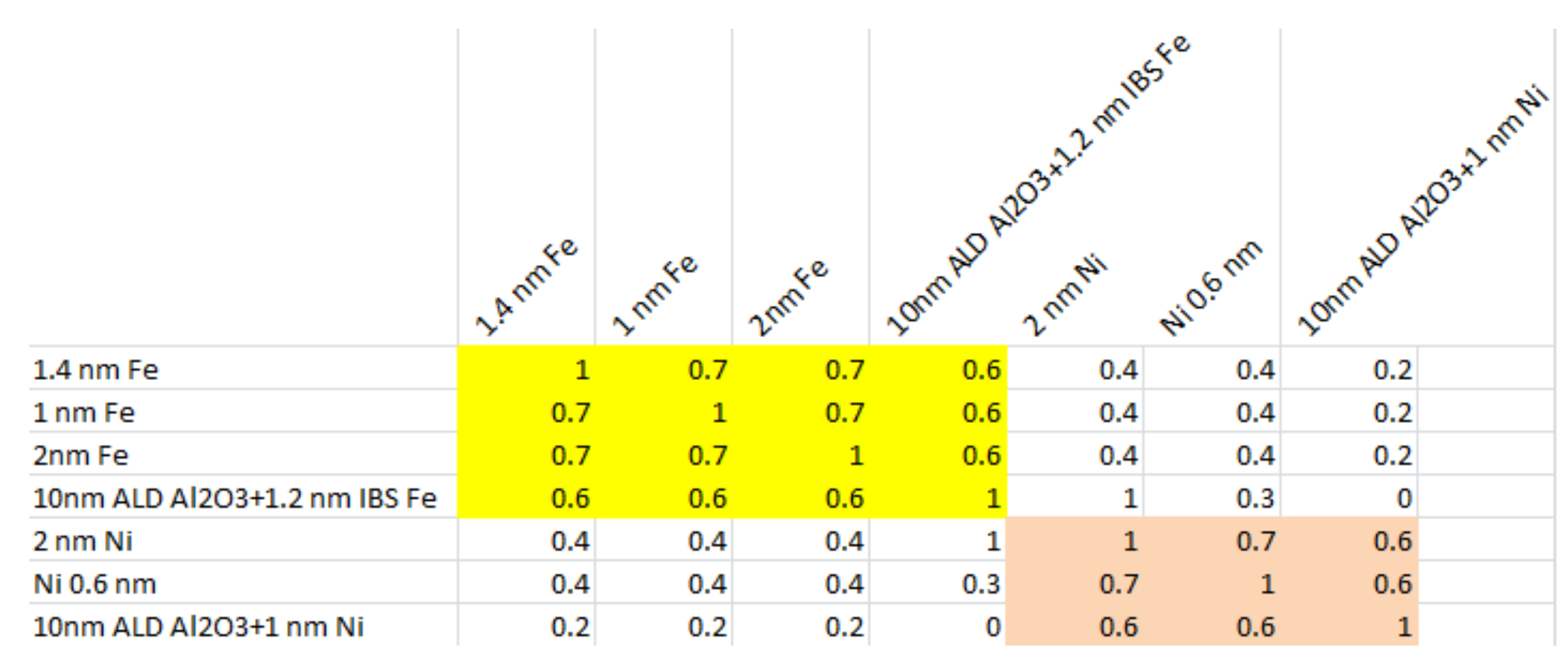}}\\
  \caption{Correlation coefficient matrix for a set of catalysts used in carbon nanotube frabrication, showing high correlations for catalysts that share nickel or iron as an element.}\label{fig:correlated}
\end{figure}

It is virtually always the case that if there is a large number of designs, then there will be similarities between the designs.  For example, imagine that we are trying to maximize the probability of growing single-walled nanotubes from a batch which is created using one of set of catalysts, some of which contain iron while the rest contain nickel.  If $x$ is the choice of catalyst, let $\mu_x$ be the unknown probability of single-walled nanotubes that would be created using catalyst $x$.  Further, let $\sigma_x$ be the standard deviation in our uncertainty in the unknown true performance $\mu_x$, and let $Cov(\mu_x,\mu_{x'})$ be the covariance which captures the behavior that if $\mu_x$ appears to be higher than we thought, then we might think that $\mu_{x'}$ will be higher as well (assuming they are positive correlated).  It is useful to write the covariance in terms of its correlation coefficient $\rho_{x,x'}$, where
\bns
Cov(\mu_x,\mu_{x'}) = \sigma_x \sigma_{x'} \rho_{x,x'}.
\ens
The correlation coefficient enjoys the property that $-1 \leq \rho_{x,x'} \leq 1$.  Figure \ref{fig:correlated} shows a correlation coefficient matrix for seven catalysts, which was estimated using the judgment of a scientist.

Correlated beliefs is an exceptionally powerful property because it allows us to generalize the results from a single experiment.  Correlations also arise when $x$ is a discretization of a continuous variable.  For example, imagine that $x$ is the concentration of a chemical.  If $x$ and $x'$ are close to each other, we would expect $\mu_x$ and $\mu_{x'}$ to be  highly correlated.  In fact, it is common to write
\bn
Cov(x,x') = \sigma_x \sigma_{x'} e^{-\beta|x-x'|}, \label{eq:exponentialcovariance}
\en
where $\beta$ is a tunable parameter that captures how quickly the covariance decays.

When we have correlated beliefs, the simple updating equations \eqref{eq:bayesmeanx} and \eqref{eq:bayesprecisionx} for independent beliefs are replaced with
\bn
\theta^{n+1}(x)    &=& \theta^n + \frac{W^{n+1} - \theta^n_x}{(\sigma^W)^2 + \Sigma^n_{xx}}\Sigma^n e_{x}, \label{eq:TransitionMu1a}\\
\Sigma^{n+1}(x) &=& \Sigma^n - \frac{\Sigma^n e_x(e_x)^T\Sigma^n}{(\sigma^W)^2  + \Sigma^n_{xx}}. \label{eq:TransitionSigma1a}
\en
(assuming we have chosen to run an experiment with controls $x$).

To illustrate, assume that we have three alternatives with  mean vector
\bns
\theta^{n} = \left[ \begin{array}{c} 20 \\ 16 \\ 22 \end{array} \right].
\ens
Assume that $\lambda^W = 9$ and that our covariance matrix $\Sigma^n$ is given by
\bns
\Sigma^n = \left[ \begin{array}{ccc}
12 & 6 & 3 \\
6  & 7 & 4 \\
3 & 4 & 15 \end{array} \right].
\ens
Assume that we choose to measure $x=3$ and observe $W^{n+1} = W^{n+1}_3 = 19$.  Applying equation \eqref{eq:TransitionMu1a}, we update the means of our beliefs using
\bns
\theta^{n+1}(3) &=& \left[ \begin{array}{c} 20 \\ 16 \\ 22 \end{array} \right] + \frac{19 - 22}{9 + 15}
\left[ \begin{array}{ccc}
12 & 6 & 3 \\
6  & 7 & 4 \\
3 & 4 & 15 \end{array} \right] \left[ \begin{array}{c} 0 \\ 0 \\ 1 \end{array} \right] \\
          &=& \left[ \begin{array}{c} 20 \\ 16 \\ 22 \end{array} \right] + \frac{-3}{24} \left[ \begin{array}{c} 3 \\ 4 \\ 15 \end{array} \right]  \\
          &=& \left[ \begin{array}{c} 19.625  \\ 15.500 \\ 20.125 \end{array} \right].
\ens
The update of the covariance matrix is computed using
\bns
\Sigma^{n+1}(3) &=& \left[ \begin{array}{ccc}
12 & 6 & 3 \\
6  & 7 & 4 \\
3 & 4 & 15 \end{array} \right] - \frac{\left[ \begin{array}{ccc}
12 & 6 & 3 \\
6  & 7 & 4 \\
3 & 4 & 15 \end{array} \right]\left[ \begin{array}{c} 0 \\ 0 \\ 1 \end{array} \right][0 ~~0 ~~1]\left[ \begin{array}{ccc}
12 & 6 & 3 \\
6  & 7 & 4 \\
3 & 4 & 15 \end{array} \right]}{9+15} \\
&=& \left[ \begin{array}{ccc}
12 & 6 & 3 \\
6  & 7 & 4 \\
3 & 4 & 15 \end{array} \right] - \frac{1}{24}\left[ \begin{array}{c} 3 \\ 4 \\ 15 \end{array} \right][3 ~~4 ~~15] \\
&=& \left[ \begin{array}{ccc}
12 & 6 & 3 \\
6  & 7 & 4 \\
3 & 4 & 15 \end{array} \right] - \frac{1}{24}\left[ \begin{array}{ccc}
9 & 12 & 45 \\
12 & 16 & 60 \\
45 & 60 & 225 \end{array} \right]  \\
&=& \left[ \begin{array}{ccc}
12 & 6 & 3 \\
6  & 7 & 4 \\
3 & 4 & 15 \end{array} \right] - \left[ \begin{array}{ccc}
0.375 & 0.500 & 1.875 \\
0.500  & 0.667 & 2.500 \\
1.875 & 2.500 & 9.375 \end{array} \right] \\
&=& \left[ \begin{array}{ccc}
11.625 & 5.500 & 1.125 \\
5.500  & 6.333 & 1.500 \\
1.125 & 1.500 & 5.625 \end{array} \right].
\ens
Notice how running experiment 3 (by this we mean a particular set of design parameters), changes our belief about all other experiments, and updates the entire covariance matrix.

These calculations are fairly easy when the number of alternatives is up to around 1,000; beyond that, the matrices become quite clumsy to work with.  However, this logic has made it possible to find high quality solutions when there are hundreds of alternatives with as few as 10 or 20 experiments (this depends on the covariance matrix).  The effect of correlated beliefs is to make the set of possible experimental designs much smaller than it seems.

\subsection{Linear (parametric) models}
There are many settings where the number of potential design decisions $x\in\Xcal$ is simply much too large to use a lookup table belief model, even with correlated beliefs.  In some settings (and this is fairly common), it is possible to replace the lookup table model with a statistical model.
\begin{figure}[tb]
  \center{\includegraphics[width=4.0in]{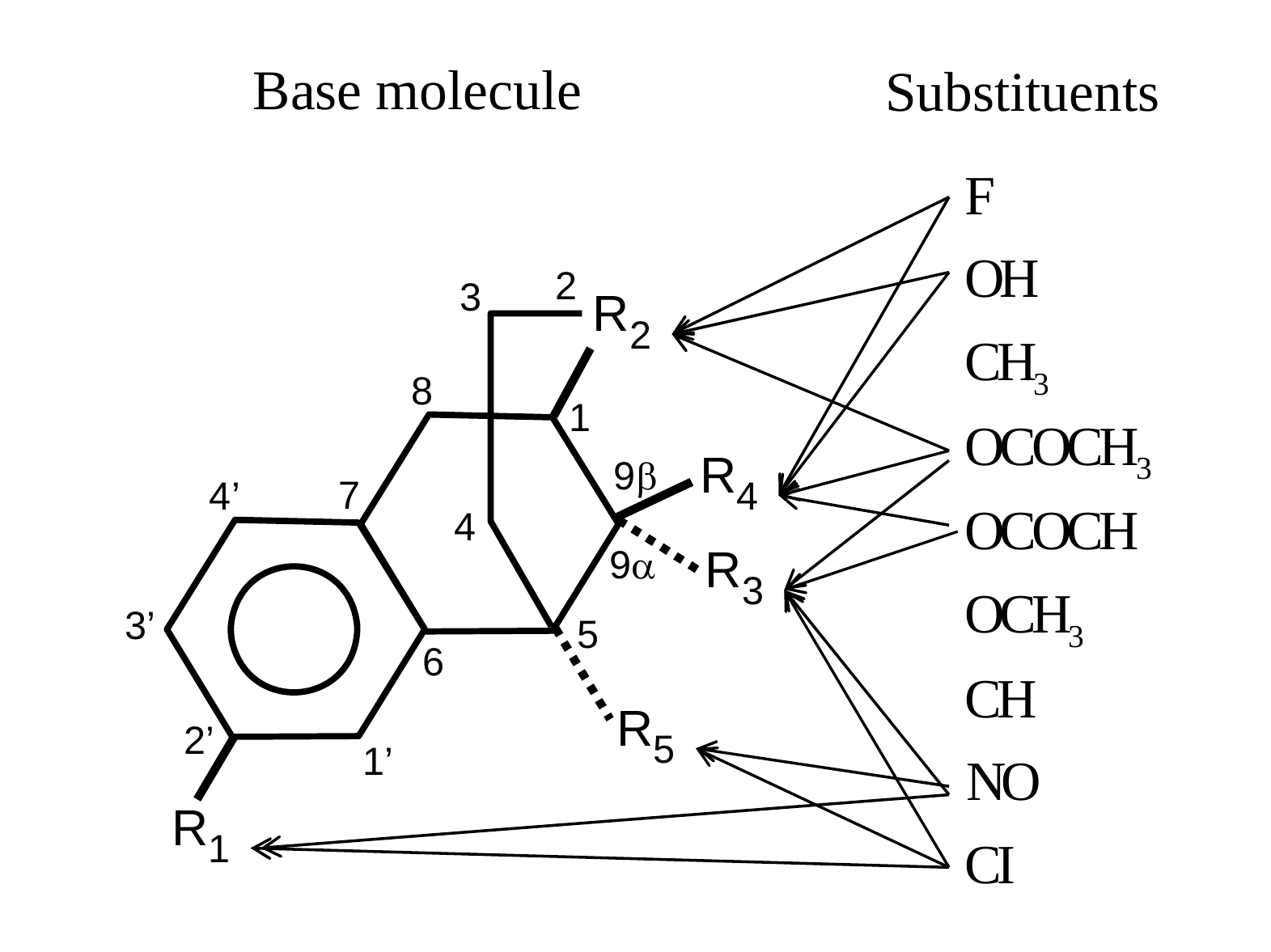}}\\
  \caption{Base molecule with a set of sites $R_1$ \ldots $R_5$ for substituents.}\label{fig:Basemoleculewsubstituents}
\end{figure}

A real application of this approach is modeling the behavior of drugs being used to kill cancer cells. Imagine that we are trying to build a molecule by attaching a substituent $i\in\Ical$ to a site $j\in\Jcal$ on the molecule.  We let $X_{ij} = 1$ if we assign the $ith$ substituent to the $jth$ site, as depicted in figure \ref{fig:Basemoleculewsubstituents}. In this case, our experiment $x$ would be given by the vector $x=(X_{ij})_{i\in\Ical,j\in\Jcal}$, which means the number of possible values of $x$ can be extremely large. We might write our response (material strength, cancer cells killed) using a linear statistical model
\bn
Y = \theta_0 + \sum_{i\in\Ical} \sum_{j\in\Jcal}\theta_{ij} X_{ij} + \epsilon. \label{eq:linearmodel}
\en
This is called a {\it linear} model because it is linear in the unknown parameters $\theta = (\theta_0, (\theta_{ij})_{ij})$.  In the science literature, this is referred to as a QSAR model (for quantitative structural activity relationship).  The variable $\epsilon$ is a random variable capturing experimental variation, where we generally assume that $\epsilon \sim N(0, \sigma^2_\epsilon)$.  The experimental variance $\epsilon^2_\epsilon$ can typically be estimated using a few experiments and updated over time. A more general form of our linear model is typically written
\bn
Y(x,\theta) = \theta_1 \phi_1(x) + \theta_2 \phi_2(x) + \ldots + \theta_F \phi_F(x) \label{eq:linearmodel2}
\en
where $(\phi_f(x))_{f\in\Fcal}$ is a set of {\it features} extracted from a (possibly complicated) set of characteristics of our experiment which we represent by $x$.  In this setting, $x$ can consist of control parameters (temperatures, pressures, densities) as well as observed features that we may not control (humidity, voltage variability).

With a linear model such as \eqref{eq:linearmodel2}, instead of having to estimate $\mu_x$ for each $x$ (which could easily number in the hundreds of thousands to millions), we just have to find the parameter vector ${\bm \theta}$, at which point we let $\mu_x = Y(x|\theta)$.  Here, the true value of ${\bm \theta}$ is a random variable (just as the truth $\mu_x$ was random when we used our lookup table representation).  As before, we expect to have an initial estimate $\theta^0$ and variance $\sigma^{2,0}$.

Since the linear model in \eqref{eq:linearmodel2} is so much more compact than a lookup table model, one might ask why you would ever do a lookup table model.  It is important to realize that we are paying a price using the linear model; it is imposing a specific structure that may not actually be true.  For example, in our model of different molecules depicted in figure \ref{fig:Basemoleculewsubstituents}, a linear model requires that we make the assumption that the marginal effect of $CH_3$ in one site has nothing to do with whether $OCH_3$ has been included in another site.  This may not be accurate, but this is one of the limitations of a linear model.

\begin{figure}[tb]
  \center{\includegraphics[width=4.0in]{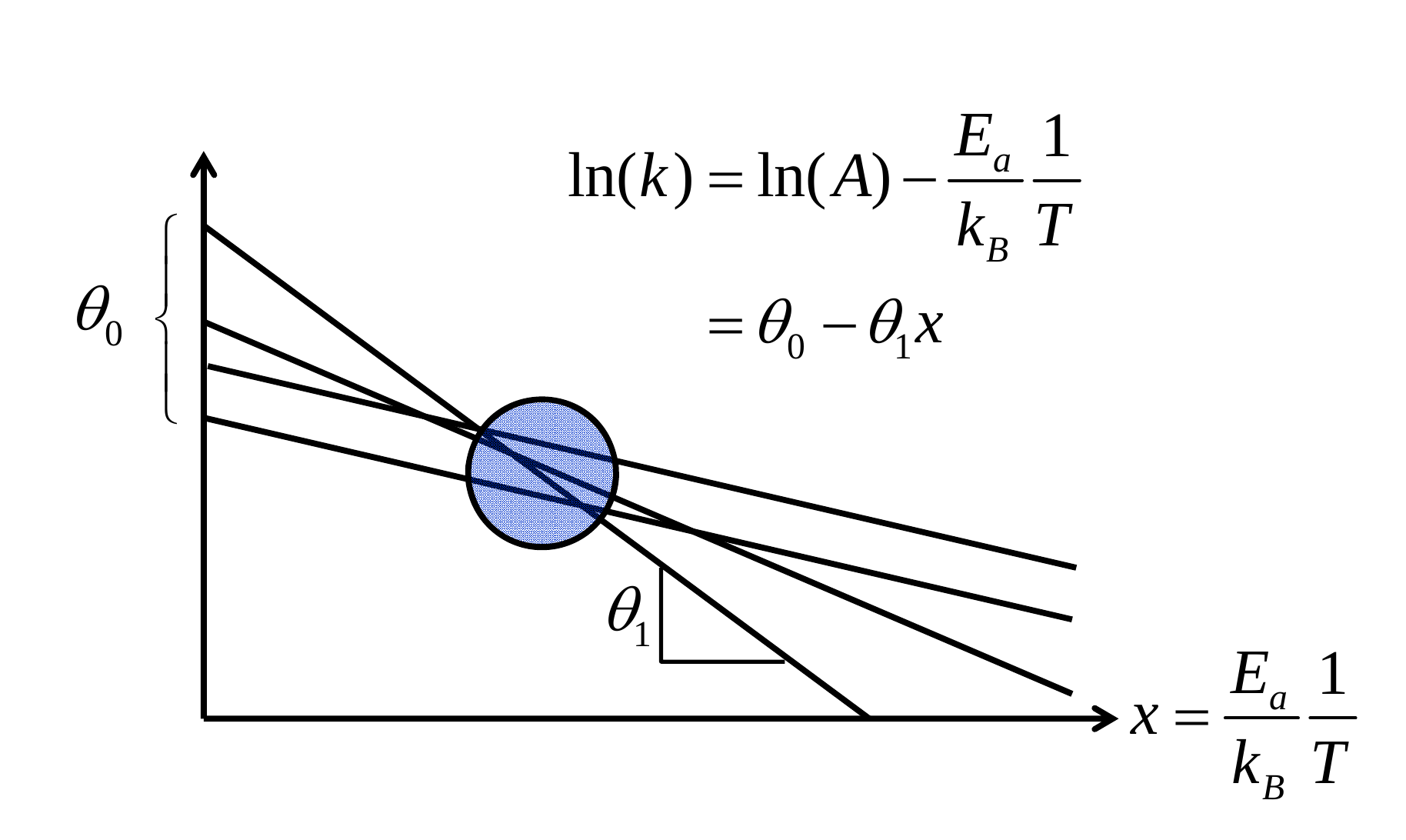}}\\
  \caption{Arrhenius plot, showing a sample of possible relationships between temperature $T$ and the reaction rate $k$.}\label{fig:correlatedlines}
\end{figure}
Transitioning to a linear belief model introduces some modeling subtleties.  We have seen that it is important to model correlations between values of $\mu_x$ and $\mu_{x'}$ when using a lookup table belief model.  With a parametric belief model, we directly capture the relationship between $\mu_x = Y(x|\theta)$ and $\mu_{x'} = Y(x'|\theta)$ since they are connected through the linear model.  Less obvious is that we still need to capture correlations between the parameters ${\bm \theta}$.  This is shown in figure \ref{fig:correlatedlines} which illustrates the variability of a random set of lines (relating reaction rate $k$ to temperature $T$) which have a tendency to move through a central region assumed by a scientist.  In addition to the pure variability of the intercept $\theta_0$ and slope $\theta_1$,  they also tend to be negatively correlated, since a higher intercept tends to be associated with a steeper slope.

If we run an experiment with experimental choice $x=x^n$ and observe outcome $W^{n+1} = Y^{n+1}$, we can update our belief using the equations
\bn
\theta^{n+1} = \theta^n + \frac{1}{\gamma^n}B^n x^n \varepsilon^{n+1}, \label{eq:recursive1}
\en
where $\varepsilon^n$ is the error given by
\bn
\varepsilon^{n+1} = y^{n+1} - Y(x|\theta^n). \label{eq:recursive2}
\en
where $Y(x|\theta^n)$ is given by \eqref{eq:linearmodel2}.  Letting $\sigma^2_\epsilon$ be the variance of $\epsilon$, we can write $\Sigma^{\theta,n} = B^n \sigma^2_\epsilon$.  The matrix $B^n$ can be updated recursively using
\bn
B^{n+1} = B^n - \frac{1}{\gamma^n} (B^n x^{n} (x^{n})^T  B^n).  \label{eq:recursive4}
\en
The scalar $\gamma^n$ is computed using
\bn
\gamma^n &=& 1+(x^n)^TB^nx^n.   \label{eq:recursive5}
\en

These equations illustrate that the process of updating a linear belief model is actually relatively simple.  The harder problem is developing the initial prior, a problem we return to below.

\subsection{Sampled nonlinear models}
\label{sec:samplednonlinear}
Assume that the outcome $Y$ of our experiment is binary, where $Y=1$ represents a success (for example, creating a single-walled nanotube, a probe binding to an RNA molecule or a successful medical treatment) while $Y=0$ is a failure.  A success can be attributed to observable state variables that we do not control (such as the humidity in the lab or the attributes of a patient such as gender or a smoking habit), as well as decisions we make (the material flux, temperature, or specific medical treatments such as drugs or surgery).  We may feel that we can write the probability of a successful outcome (a single-walled nanotube, or successful treatment of a patient) using a logistic function, which is written as
\bn
P(Y=1|s,x,\theta) = \frac{e^{\theta_1 s_1 + \theta_2 s_2 + \theta_3 x_1 + \theta_4 x_2 + \ldots}}{1+e^{\theta_1 s_1 + \theta_2 s_2 + \theta_3 x_1 + \theta_4 x_2 + \ldots}}. \label{eq:logisticregression}
\en
For example, $s_1$ might be the humidity in the lab while $s_2$ is the outdoor temperature, while $x_1$ could be the choice of catalyst while $x_2$ is the material flux.  In a medical setting, $s_1$ and $s_2$ could be patient attributes, while $x_1$ and $x_2$ could be medical decisions (tests and drugs).

\begin{figure}[tb]
  \begin{center}
  \begin{tabular}{cc}
  \includegraphics[width=3.0in]{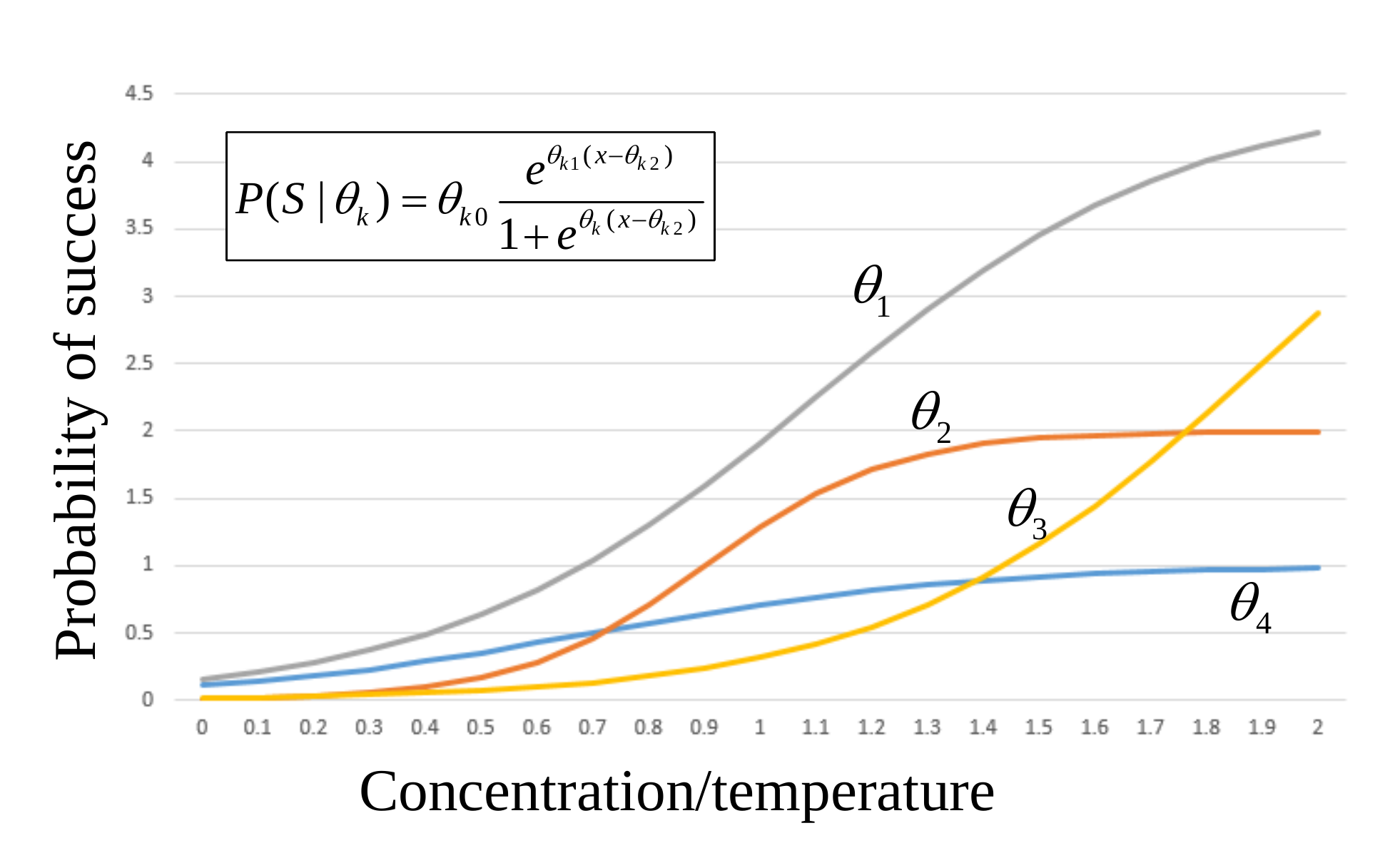} & \includegraphics[width=3.0in]{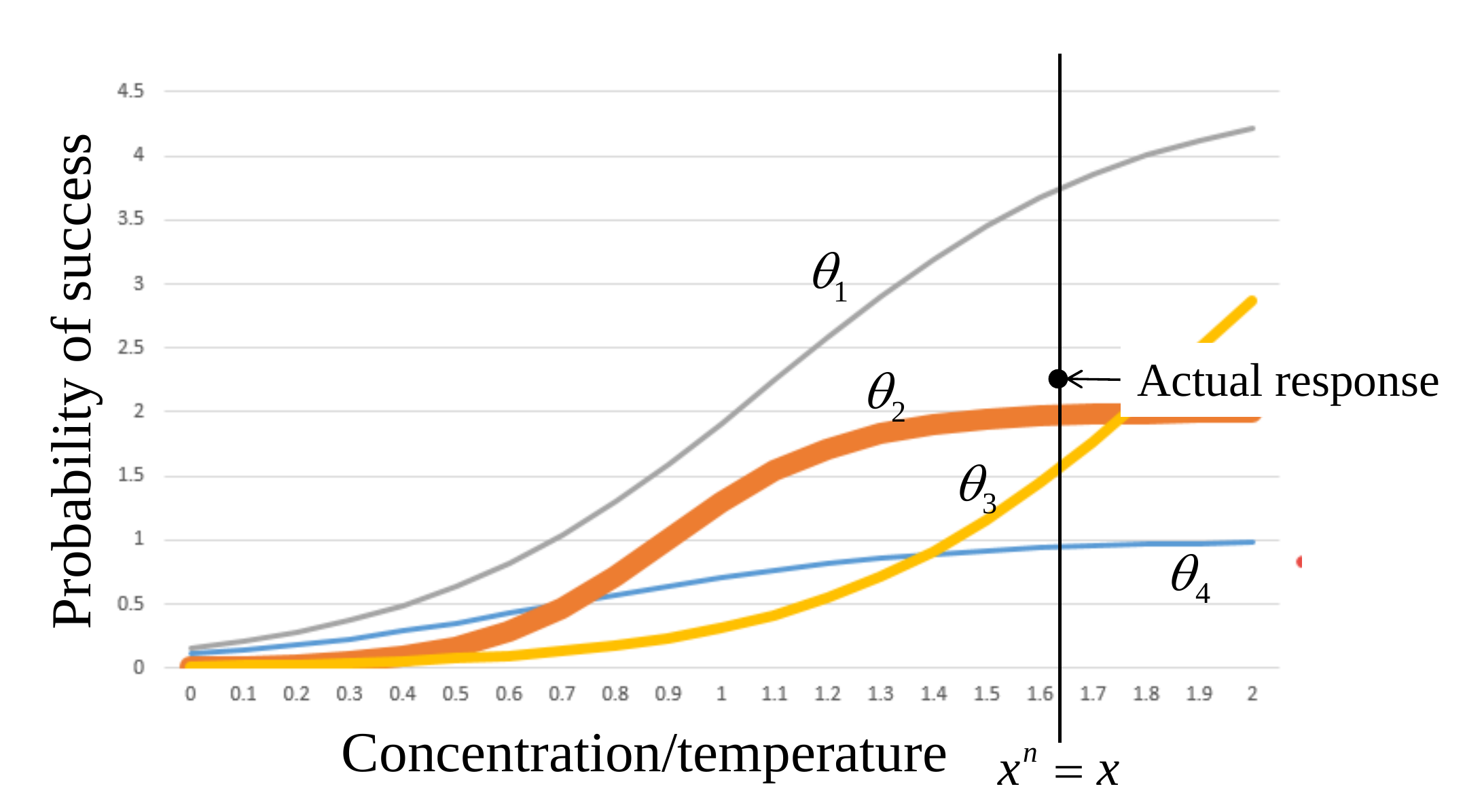} \\
                            (a)                      &                           (b)
  \end{tabular}
  \end{center}
  \caption{Sample of possible relationships between temperature $T$ and the reaction rate $k$.}\label{fig:sampledbelief}
\end{figure}

Up to now, we have represented the uncertainty in parameters (whether in a lookup table model or a linear model) using a multivariate normal distribution.  Now, we are going to use a much simpler strategy known as a sampled belief model, where we assume that the unknown parameter vector ${\bm \theta}$ might take on one of a set of specific values $\theta_1, \ldots, \theta_K$, where each vector $\theta_k = (\theta_{k1}, \theta_{k2}, \ldots, \theta_{kI})$ consists of the parameters required to specify the function in \eqref{eq:logisticregression}.  Figure \ref{fig:sampledbelief}(a) illustrates our probability model \eqref{eq:logisticregression} for four different values of $\theta$.

When we used a multivariate normal distribution, we started with estimates such as $\theta^0_x$ and a variance $\sigma^{2,0}_x$, and then assumed that the truth $\mu_x$ (for a lookup table belief model) or $f(x,{\bm \theta})$ (in a parametric model) was normally distributed.  With a sampled belief model, we assume that we have a discrete distribution $(p^0_1, \ldots, p^0_K)$ of the possible values $(\theta_1, \ldots, \theta_K)$ that the true vector ${\bm \theta}$ might be, where $p^0_k = \mbox{Prob}[{\bm \theta} = \theta_k]$.  We might start by assuming that the different values $\theta_k$ are equally likely.

Just as we demonstrated the updating equations for our first two belief models, we now demonstrate how to update a sampled belief model using a simple relationship known as Bayes theorem, which is fundamental to any information collecting process.  If we have a random event $A$ (for example, whether the experiment was successful or not), and a random event $B$ (in our setting, this will correspond to which value of ${\bm \theta}$ is true), we start with the basic relationship
\bns
P(A,B) &=& P(A|B) P(B) = P(B|A) P(A),
\ens
which says that the joint probability of events $A$ and $B$ is equal to the conditional probability of $A$ given that $B$ has happened times the probability that $B$ happens, which is also equal to the probability that $B$ happens given that $A$ has happened, times the probability of $A$.  This relationship quickly produces
\bn
P(B|A) = \frac{P(A|B) P(B)}{P(A)}, \label{eq:bayestheorem}
\en
which is Bayes theorem.  Now replace $A$ with the random variable $Y$ that captures whether or not we observed a success, and let $B$ represent one of the $K$ values of $\theta_k$.  The probability $P(B)$ is the prior (if we have just finished the $nth$ experiment, this would be the probability vector $p^n = (p^n_1, \ldots, p^n_K)$).  The conditional probability $P(A|B) = P(Y|\theta)$ is calculated directly from equation \eqref{eq:logisticregression}, since we get to assume we know what $\theta$ is.  This allows us to write
\bn
p^{n+1}_k(Y^{n+1}) &=& P({\bm \theta}=\theta_k|Y^{n+1}=1) \nonumber \\
                   &=& \frac{P(Y^{n+1}=1|\theta_k) p^n_k}{P(Y^{n+1}=1)} \label{eq:bayesupdate}
\en
where $P(Y^{n+1} = 1) = \sum_{k=1}^K P(Y^{n+1}=1|\theta_k) p^n_k$ is just the probability that $Y^{n+1} = 1$ from equation \eqref{eq:logisticregression} when averaged over all possible values of $\theta$.  Finally, we get the $P(A) = P(Y^{n+1}=1)$ in the denominator of \eqref{eq:bayestheorem} by just averaging over the different values of $\theta$ using
\bns
P(Y^{n+1}=1) = \sum_{k=1}^K P(Y^{n+1}=1|\theta_k) p^n_k.
\ens

Bayes theorem is of fundamental importance in any arena (such as laboratory experimentation) that involves collecting information.  This is the fundamental equation for making the transition from our prior distribution ($P(B)$ above, or $P({\bm \theta} = \theta_k)$), which is our distribution of belief before we observe new information, and the posterior distribution $P(B|A) =  P({\bm \theta} = \theta_k|Y^{n+1})$.

Sampled belief models are particularly nice to work with because the uncertainty is expressed in such a simple way.  As we get into certain classes of policies, we are going to find that sampled belief models offer some fairly nice computational advantages for certain types of policies, especially when the belief model is nonlinear in the unknown parameter $\theta$, as it is in \eqref{eq:logisticregression} (this is common with many physical models).

\subsection{Creating priors}
All experimental projects have to start with the first experiment, which requires that you make your first decision before you have any data.  What you know at this point is called your {\it prior}.  This introduces the question: how do you create your prior?  There are a number of strategies, some of which include:
\begin{itemize}
\item Knowledge of physics or chemistry - The science of a problem may provide an initial indication of what is possible.
\item Literature review - Scan the vast literature for the experiences of other scientists who worked on similar problems.
\item Numerical simulations - Numerical simulators often provide initial approximations of the properties of a material or the kinetics of the chemistry.
\item Subjective judgment - Drawing on extensive experience (supported by an understanding of the chemistry), a scientist may be able to make reasoned guesses.
\item Exploratory simulations - A scientist may run a few exploratory experiments just to get an initial sense of how an experiment responses to certain inputs.
\end{itemize}

\begin{figure}[tb]
  \begin{center}
  \begin{tabular}{cc}
  \includegraphics[width=3.0in]{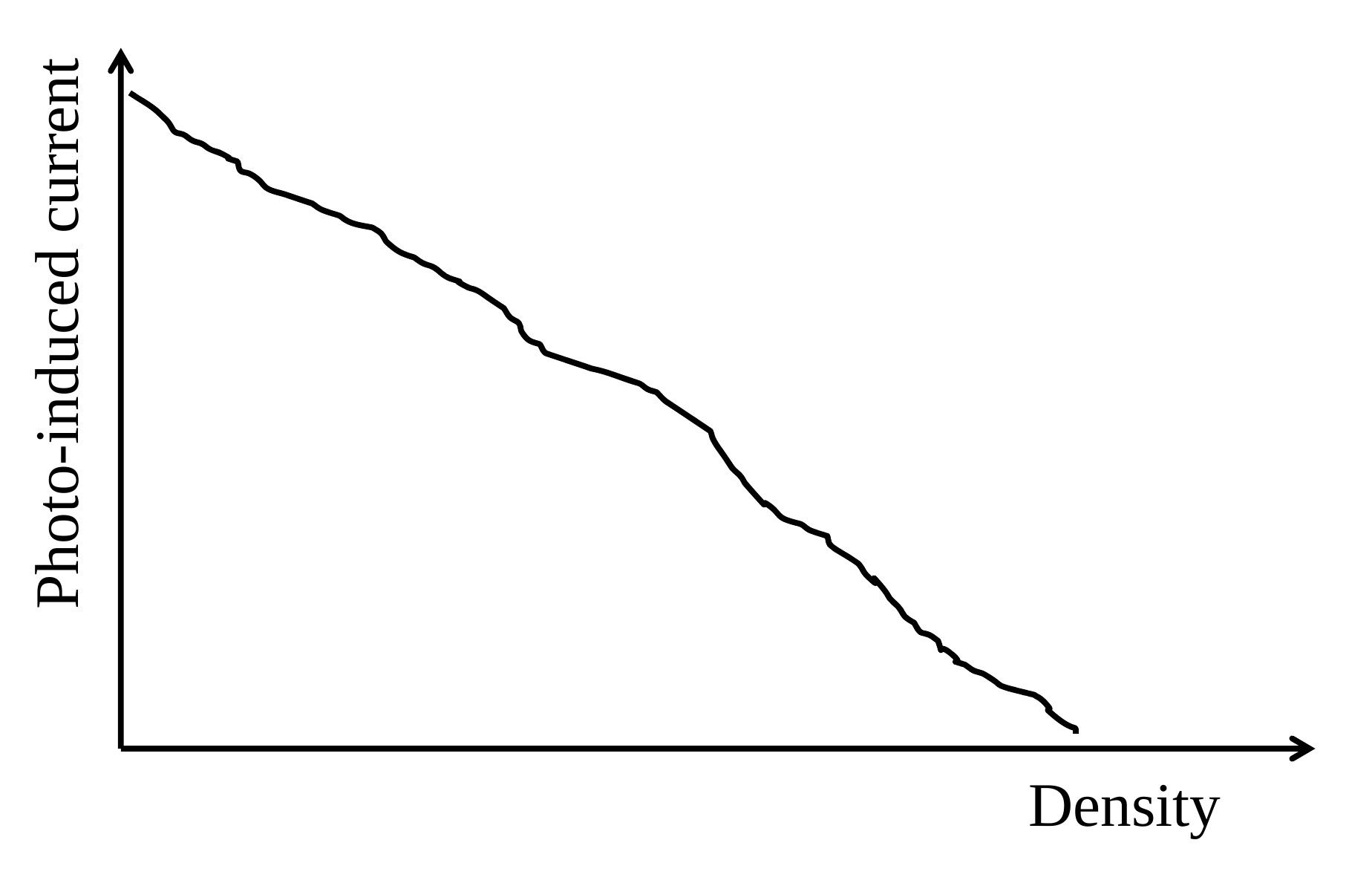} & \includegraphics[width=3.0in]{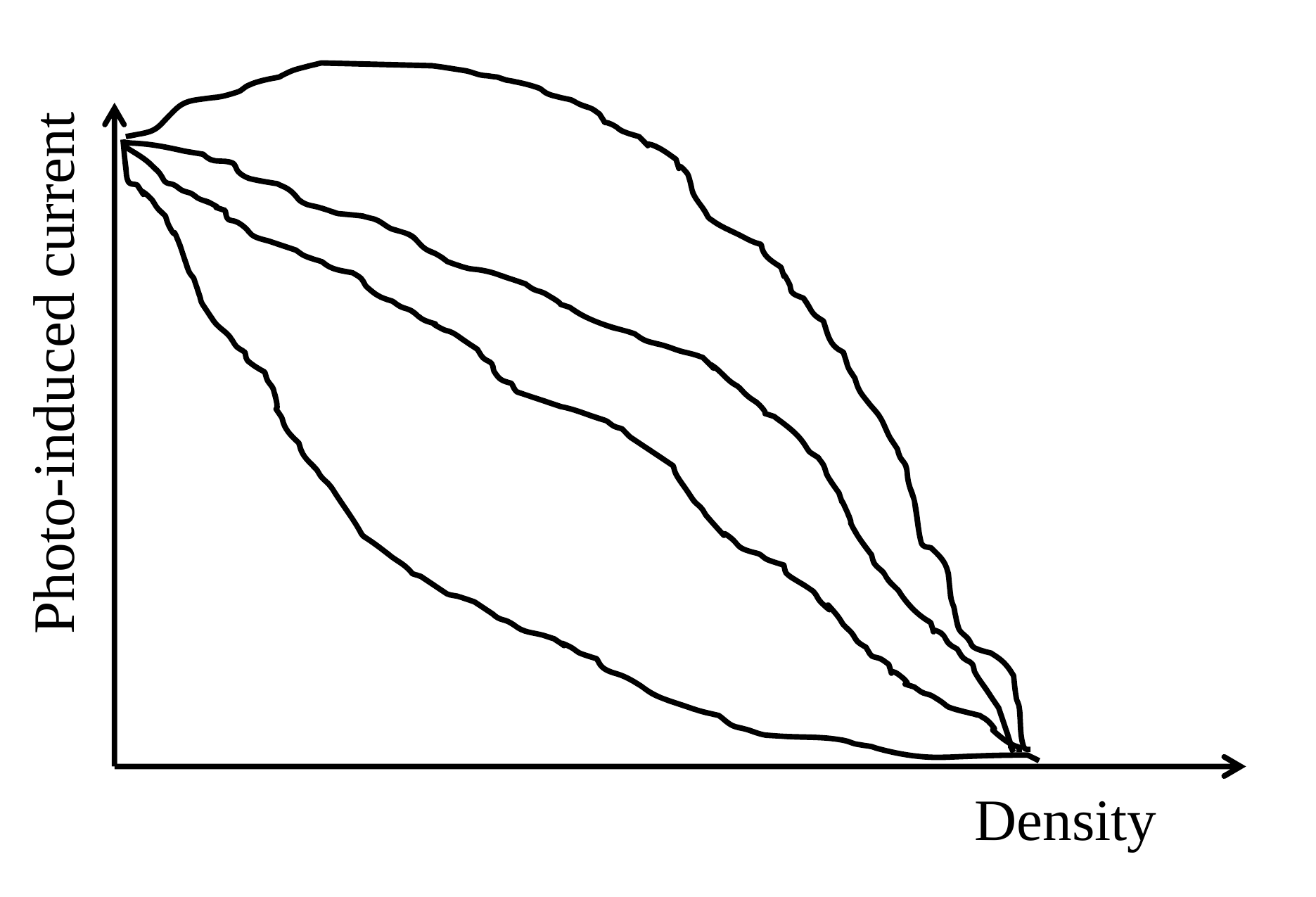} \\
                            (a)                      &                           (b)
  \end{tabular}
  \end{center}
  \caption{Point estimate of a relationship (a), sampled distributional estimate of a relationship (b)}\label{fig:priorline}
\end{figure}

Figure \ref{fig:priorline} depicts the process that actually occurred with one team.  When asked for their best guess of the relationship between a photo-induced current and the density of nanoparticles attached to a substrate of a photoelectric device, the team drew the line in figure \ref{fig:priorline}(a).  We then challenged them to guess what the relationship {\it might} be, and this produced the diagram in \ref{fig:priorline}(b).  The difference is critical.  Figure \ref{fig:priorline}(a) suggests a perfect understanding of the relationship, although of course this would not be the case.

Figure \ref{fig:priorline}(b) captures the uncertainty in the relationship, but note that this uncertainty arises in a very specific way.  Apparently the scientist had some reason to believe that all the curves started and ended at the same point.  This is an important reason why it is essential to indicate the uncertainty in your belief.

An approach with a lookup table belief model is to make a best guess (the point estimate) and then specify error bars to capture the tails of the distribution (flip back to figure \ref{fig:basiclearning}).  Recall that the correlation coefficient matrix capturing the relationship between different catalysts given in figure \ref{fig:correlated} was specified by a scientist using subjective judgment.

The key points to remember in specifying a prior are:
\begin{itemize}
\item Use whatever you know.
\item Do your best to capture problem structure, whether it is through an analytic function (with unknown parameters) or specific behaviors (as in figure \ref{fig:priorline}(b)).
\item It is less important to have the right guess than it is to be honest about the uncertainty in your guess.  It is {\it very} important that the truth be within your spread of uncertainty.
\end{itemize}

\section{Designing Policies}
\label{sec:policies}
We now address the challenge of designing effective policies that will help you achieve your objectives as quickly as possible, with minimum risk.  To do this, we have to make the right experimental decisions.  We cannot anticipate what we are going to learn, so we cannot plan what decision we will make in advance.  Instead, we have to design effective {\it policies}.

We have used this term before, but what exactly does it mean?  A policy is a method (or rule) for making a decision.  But how do we design effective policies?  It turns out that there are only two broad strategies for designing policies, each of which can be further divided into two more classes, creating four classes altogether.  All four are quite popular in learning problems, but not all are well suited to the setting of expensive experiments.  The complete list is given by:
\begin{description}
\item[Policy search] - These are rules (or functions) that have to be tuned using a simulator.  These policies divide into two subclasses:
    \begin{itemize}
    \item[1)] Policy function approximations (PFAs) - This covers any rule (this might be simple ``if ... then ... else'' rules, or analytical functions) that specifies what experiment to run given what we know now.  PFAs represent the most basic decision-making process used by people in day to day decision making.
    \item[2)] Cost function approximations (CFAs) - Here we create some kind of approximate cost function (call it $\Cbar(S,x)$), and then choose the experiment $x$ that minimizes this cost.
    \end{itemize}
\item[Policies based on lookahead approximations] - These are functions that make the best decision now by approximating the impact on the future.
    \begin{itemize}
    \item[3)] Policies based on value function approximations (VFAs) - If we are in a particular state $S$ (including physical state and belief state), and an experiment $x$ takes us to a new state $S'$ (perhaps a new physical state because we had to setup a piece of equipment, as well as a new belief state because of the information we learned), we can develop a function $\Vbar(S'(S,x))$ to approximate the value of starting in state $S'$ and proceeding until the end of some horizon, and use this to help make the best decision now.
    \item[4)] Policies based on lookahead approximations - Just as you might make a move in chess, hold your finger on the piece and think of the steps that might happen in the future, lookahead approximations try to plan decisions over some horizon to make the best decision now.  We can roughly divide these strategies into two classes:
        \begin{itemize}
        \item[4a)] One-step lookahead - There are several effective strategies that make decisions now by looking just one step out.
        \item[4b)] Multi-step lookahead - Here we plan several (sometimes many) steps into the future.
        \end{itemize}
    \end{itemize}
\end{description}
We emphasize that {\it all} of these strategies are used in learning problems, although not all are effective in the setting of expensive laboratory experiments.  However, we will emphasize that this framework covers all classes of policies, which means it includes whatever a lab is using right now (albeit informally).

One issue that all policies strive to address is the tradeoff between {\it exploration}, where we focus on learning the truth about a problem, and {\it exploitation}, where we focus on trying to maximize some objective. The best policies address this tradeoff explicitly, and the emphasis on each typically changes over the course of a set of experiments.  For example, scientists may start a sequence of experiments by just doing some random exploration, without regard to achieving any particular objective.  As the experiments progress, there is increasing interest in doing experiments which are viewed as successful.

Below we provide a bit more detail on the different types of policies, highlighting in the process which are likely to be more effective in a laboratory science setting.  However, all of the policies reviewed in this section have attracted attention for different classes of learning problems.

\subsection{Policy search}
Policy search refers to the process of tuning a rule (or function) so that the decisions it makes work well over time.  We identify two classes of rules which we refer to as {\it policy function approximations} (or PFAs) and parametric {\it cost function approximations} (or CFAs).

\subsubsection{Policy function approximations}
A PFA is any rule (or function) that specifies a decision given our state (state of knowledge, as well as physical state), without doing any form of minimization or maximization (all remaining policies include some sort of minimization or maximization within the function).  Some examples are:
\begin{description}
\item[Pure exploration] - Imagine that we have 100 different possible experiments we might run.  Pure exploration simply picks one of these at random.
\item[Boltzmann exploration] - Let $U(S^n,x)$ be the estimated utility of running experiment with controls $x$ given that our state of knowledge is given by $S^n$.  Again imagine that we have a discrete set of possible experiments $\Xcal = \{1, \ldots, M\}$.  Now pick experiment $x$ with probability
    \bn
    P(x|S^n,\beta) = \frac{e^{\beta U(S^n,x)}}{\sum_{x'\in\Xcal}e^{\beta U(S^n,x')}}, \label{eq:boltzmann}
    \en
    where $\beta$ is a tunable parameter.  If $\beta = 0$, then we choose $x$ at random out of $\Xcal$.  As $\beta$ increases, we tend toward picking the experiment with the highest utility with probability 1.
\item[Continuous function approximation] - Imagine that we need to specify the temperature $T^n$ of a process (for the $nth$ experiment), but that we feel that the best temperature also depends on the humidity $H^n$ (which is our state variable).  Noting that temperature is our decision variable (that we have been calling $x$), we propose to use a policy $X^\pi(S^n|\theta)$ which determines $x^n = T^n$ given the state $S^n = (H^n,(\theta^n,\Sigma^{\theta,n}))$, using
    \bns
    X^\pi(S^n|\theta) = \theta_0 + \theta_1 H^n + \theta_2 (H^n)^2.
    \ens
    As we did with our linear model, we assume that the true $\theta$ is multivariate normal with mean $\theta^n$ and covariance matrix $\Sigma^n$.  Once again, we have to tune the vector $\theta$ in a simulator.
\end{description}
Equation \eqref{eq:boltzmann} is known as a Boltzmann distribution (also known as Gibbs sampling).  One challenge here is that we need to tune $\theta$, which is the process we refer to as policy search.  To do this, we would have to set up a simulator that simulates the process of learning and discovering the best experimental settings.

\subsubsection{Cost function approximations}
Cost functions approximations work similarly to PFAs, with the exception that we imbed the approximating function inside a min or max operator.  Some examples are:
\begin{description}
  \item[Interval estimation] - Again assume that the set of experimental decisions is discrete, where $\theta^n_x$ is our current estimate of the performance of control settings $x$, and let $\sigma^n_x$ be the standard deviation of our estimate $\theta^n_x$.  The interval estimation policy is given by
      \bn
      X^{IE}(S^n|\theta^{IE}) = \argmax_{x\in\Xcal} \big(\theta^n_x + \theta^{IE} \sigma^n_x\big).  \label{eq:intervalestimation}
      \en
      The best way to understand interval estimation is to flip back to figure \ref{fig:basiclearning}, and imagine picking off, say, the 95th percentile of each of those distributions.  Now, pick the experiment $x$ which has the highest upper tail.  Interval estimation is a form of cost function approximation because we make up a ``cost function'' which is given by $\theta^n_x + \theta \sigma^n_x$ where we have to tune $\theta$.  A nice property of interval estimation is that instead of going through a complex tuning process, we can get quite good performance picking a value such as $\theta = 2$, which means evaluating each alternative at two standard deviations above its point estimate.
  \item[Upper confidence bounding] - A popular class of policies in the computer science community take the general form
      \bn
      X^{UCB}(S^n|\theta^{UCB}) = \argmax_{x\in\Xcal} \big(\theta^n_x + \theta^{UCB} \sqrt{\frac{\log{n}}{N^n_x}}\big), \label{eq:ucb}
      \en
      where $N^n_x$ is the number of times we have tried experiment with controls $x$.  The square root term in \eqref{eq:ucb} roughly plays the role of the standard deviation in \eqref{eq:intervalestimation}.  There are many variants of upper confidence bounding policies which enjoy nice theoretical bounds (in these settings, the tunable parameter $\theta^{UCB}$ is replaced by specific coefficients, but in practice, these have to be tuned to produce good results).  See \cite{Bubeck2012} for an advanced introduction to the theoretical analysis of these policies.
  \item[Pure exploitation] - Imagine that we have a utility function $U(S^n,x)$.  A pure exploitation policy simply picks the one with the highest utility, which we write as
      \bn
      X^{Eplt}(S^n) = \argmax_{x\in\Xcal} U(S^n,x). \label{eq:pureexploitation}
      \en
      This is a special case of a cost function approximation because there is nothing to tune.
  \item[Cost function correction] - Now imagine that we are going to modify the cost function in \eqref{eq:pureexploitation} to give us
      \bn
      X^{LCFA}(S^n|\theta) = \argmax_{x\in\Xcal} \big(U(S^n,x) + \theta_1 x_1 + \theta_2 x^2 + \theta_3 S^nx\big), \label{eq:LCFA}
      \en
      where we are assuming that $S^n$ and $x$ are scalar.  We can use any linear combination of functions of the state and decision to create a general parametric cost function approximation using
      \bn
      X^{PCFA}(S^n|\theta) = \argmax_{x\in\Xcal} \big(U(S^n,x) + \sum_f \theta_f \phi_f(S^n,x). \label{eq:PCFA}
      \en
\end{description}
Through extensive experimentation, we have found interval estimation to be an unusually effective policy, but we have also found that while conventional wisdom allows us to set $\theta=2$, other values of $\theta$ can produce much better results (but it depends on the characteristics of the problem).  Also, in the world of very finite experimental budgets, the tradeoff between exploitation (that is, maximizing $U(S^n,x)$ and exploration (doing experiments where there is the greatest uncertainty, captured by $\sigma^n_x$), changes as we get close to the end of our budget.  By contrast, the two cost function approximations, LCFA and PCFA, would require careful tuning in a simulator.

\subsection{Policies based on lookahead approximations}
Policies based on lookahead approximations are so natural we do not always recognize when we are using them.  Each time we use our navigation system in our car, we are using a lookahead approximation because we compute a shortest path all the way to the destination, even though we may have to deviate from this path as events arise. Also, if you have ever played chess by moving a piece and keeping your finger on it while you think about what might happen next, you are using a lookahead policy.

A scientist is using a lookahead policy if he/she is thinking about an experiment which requires, for example, setting up a machine to work with a particular material, after which it makes sense to do a series of experiments with that material.  In fact, it is hard not to use a lookahead policy whenever there is a physical state (such as setting up the machine).

There are two strategies, both widely used for sequential decision problems, for designing policies that involve some form of lookahead: indirect lookahead policies that use some sort of approximation for the value of being in a state (physical as well as state of knowledge), and direct lookahead policies that plan one or more steps out before making a decision now.

\subsubsection{Indirect lookahead using a value function approximation}
Imagine that we are thinking about running a series of experiments to test different catalysts.  Each time we set up a machine to test a catalyst, we are then going to do a series of experiments tuning other parameters such as the density of particles, material flux, and temperature of the process.  However, rather than explicitly planning all these experiments in advance before deciding which catalyst to test, we simply use a rough approximation of the value of initiating a series of experiments for each catalyst.

We refer to these ``rough approximations'' as value function approximations, which we write as $\Vbar(S^x)$, where $S^x$ (known as a ``post-decision state'') is the state after we have run experiment using controls $x$ (in our example, this would mean setting up a machine to handle a particular catalyst), but before we have run the actual experiment.

Value function approximations are at the foundation of entire fields with names such as approximate dynamic programming (\cite{PowellADP2011}, \cite{BertsekasADP2012}) and reinforcement learning (\cite{SuBa98} is the bible of this community).  However, the methods are more complex, and as of this writing have not proven to be as useful as other strategies for learning problems.

\subsubsection{Direct lookahead}
A direct lookahead policy takes what we know now (captured by the state $S^n$), considers running a particular experiment with controls $x$, and then thinks about what might happen in the future (both decisions that we might make, and outcomes we might learn).  We start by illustrating multistep lookahead policies (which are easy to illustrate but much harder to execute), and then close by describing a powerful class of policies that are based on a single step lookahead.

\begin{figure}[tb]
  \center{\includegraphics[width=5.5in]{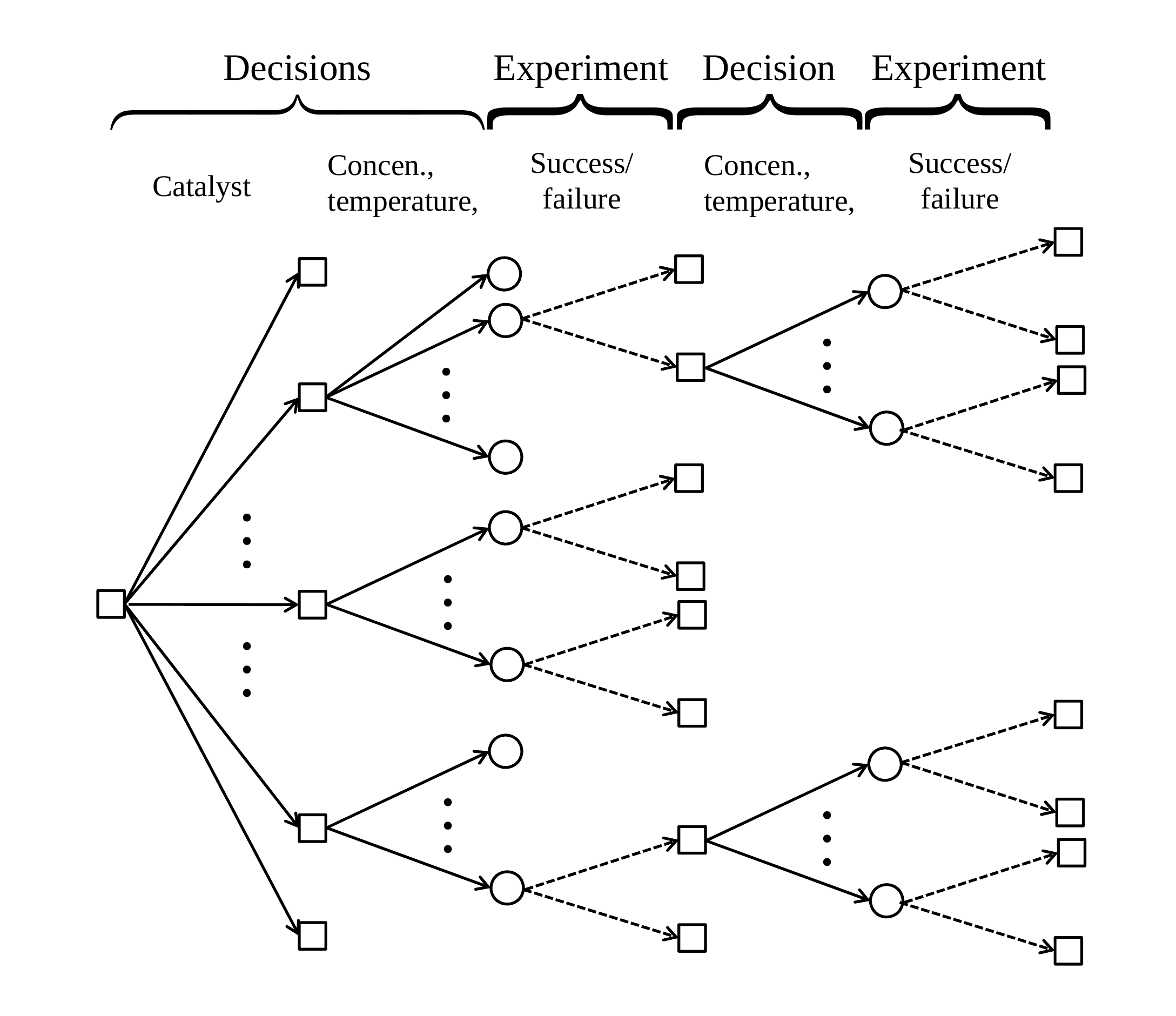}}\\
  \caption{A sequence of decisions and experimental outcomes modeled as a decision tree.}\label{fig:decisiontree}
\end{figure}

{\noindent \bf Multistep lookahead}\\
A multistep lookahead policy creates a complete decision tree of actions and experimental outcomes.  Figure \ref{fig:decisiontree} illustrates a problem which starts by choosing a catalyst, then models a sequence of experiments that involve testing different concentrations and temperatures, followed by an experimental outcome, followed by a new experiment, and so on.  In effect, the decision tree is enumerating everything that might happen, after which we can roll back to determine which catalyst we should test now.  When constructing the tree, we have to capture the process of updating belief models which we have illustrated above.

The strength of decision trees is their ability to handle a wide range of issues that arise in real experimental settings.  For example, we can model the setup of physical equipment, as well as the fact that some experiments take more time than others.  We can also impose budget constraints on the number of experiments, the total time, and the total cost.

The problem with decision trees, however, is that they can quickly explode in size. Imagine that we have 100 different experimental settings, after which we observe 10 different experimental outcomes.  If we want to think 5 experiments into the future, we will be generating $(10^{100} \times 10^{10})^5 = 10^{15}$ different pathways.  There are techniques to help overcome this complexity, but a different question is whether we actually need the capabilities (and computational overhead) that comes with generating and solving a full decision tree.  We need to remember, for example, that the PFAs and CFAs reviewed above can work very well.  In particular, there is a particularly powerful class of policies based on one-step lookahead models, which we review next.

{\noindent \bf One-step lookahead}\\
Many learning problems are particularly well suited to policies that do nothing more than look one step into the future.  These include:
\begin{description}
\item[Thompson sampling] - Assume we are using a lookup table belief model where our belief about the true performance of our process when we use controls $x$ is given by $\mu_x \sim N(\theta^n_x, \sigma^{2,n}_x)$.  Now, for each controls $x$, randomly sample from this distribution and call the result ${\hat \theta}^{n+1}_x$, which we can view as a sampled estimate of what might happen in the $n+1st$ experiment. We then find the largest ${\hat \theta}^{n+1}_x$ and let $x^n$ be the corresponding design.  Thompson sampling has become very popular in the internet community, but would be unlikely to work well in the setting of expensive laboratory experiments.  However, it could be effective in the more fast-paced setting of robotic scientists.
\item[Bayes greedy] - Imagine that we have some function $F(x|\theta)$ that is nonlinear in the unknown parameters $\theta$ (see, for example, our success/failure model \eqref{eq:logisticregression}).  One way of finding $x$ is to let $\theta = \theta^n$ (that is, set it equal to its best estimate), and then find $x$ that maximizes $F(x|\theta^n)$.  The problem is that this ignores the uncertainty in our belief about $\theta$.  Assume for simplicity that $\theta$ might be one of $\theta_1, \ldots, \theta_K$ with probabilities $p^n_1, \ldots, p^n_K$, where ${\bar \theta}^n = \sum_{k=1}^K p^n_k\theta_k$.  A simple greedy policy would solve
    \bns
    x^n = \argmax_x F(x|{\bar \theta}^n).
    \ens
    This is like assuming that the true value for ${\bm \theta}$ is equal to the expectation based on their current set of probabilities.  Bayes greedy, on the other hand, averages the {\it function} across the different possible values $F(x|\theta_1), \ldots, F(x|\theta_K)$, and then finds the value $x$ that solves
    \bn
    x^n = \argmax_x \sum_{k=1}^K p^n_k F(x|\theta_k).
    \en
    Thus, Bayes greedy is simply maximizing the expected value of what might happen in the next experiment.
\item[Value of information] - This is a powerful set of policies that is particularly well suited to expensive laboratory experiments.  In a nutshell, value of information policies look at the ability of an experiment to make better decisions, leading to better results.  Policies in this class go under names such as sequential kriging ((\cite{Huang2006}, \cite{PoRy2012}[Section 16.2]), and expected improvement (EI) (\cite{Jones1998}, \cite{PoRy2012}[Chapter 5]), but we are going to focus on the more precise version that has been developed under the name {\it knowledge gradient}, which is described in depth in the next section.
\end{description}

\subsection{Remarks}
Recognizing that all of these policies are actively used in different learning problems, it is useful to discuss which are best suited to applications in laboratory settings.  We start by noting that the policies that have to be tuned via policy search (policy function approximations and parametric cost function approximations) need a process for tuning.  This is not possible in a real laboratory setting, but it may be possible to create a simulator that can be used for tuning.  An attractive feature of PFAs and CFAs is that once tuned, they are very simple to implement.

We have considerable experience using value function approximations for a wide range of complex resource allocation problems \cite{PowellADP2011}, but as of this writing, this class of policies have yet to prove themselves for learning problems.  The difficulty is that value function approximations can be difficult to approximate, and simpler policies seem to work quite well.

We anticipate that multistep lookahead policies (using some form of decision tree) may find applications in laboratory settings that combine learning with handling a physical state (such as setting up a machine for a particular class of experiment), and some learning problems with very small budgets.  Again, as of this writing we are unaware of any interest in this approach for experimental learning problems (but stay tuned!).

Based on our own experience and the attention from the literature, the one-step lookahead policies (Thompson sampling, Bayes-greedy and the knowledge gradient) seem to have attracted considerable interest from the research literature (and scientific community).  Our concern with Bayes greedy is that it does not recognize the value of doing experiments where there is greater uncertainty (something that is clearly evident in the interval estimation policy in \eqref{eq:intervalestimation}).  Thompson sampling indirectly captures the value of uncertainty because experiments that exhibit high uncertainty are more likely to be sampled.  However, this policy is best suited for settings with large experimental budgets, since you cannot decide on an experiment that may take several days (and many thousands of dollars) based on Monte Carlo sampling.

So which policy is best?  To help with this question, we have created a new environment called MOLTE (Modular Optimal Learning Testing Environment) which is a Matlab-based testing system for comparing a wide range of experimental policies (using either online or offline objectives) on a library of different problems.  Each policy is implemented as its own Matlab function, and each test problem is also implemented in its own Matlab function.  This makes it easy for researchers to introduce new policies, and new problems, just by following some simple software protocols.  The software and a users manual can be downloaded from: \\
\indent \url{http://castlelab.princeton.edu/software.htm#molte}.

We have worked extensively with a wide range of policies.  We have found that the knowledge gradient is particularly well suited to expensive experimentation.  The next section explains why.

\section{The knowledge gradient}
\label{sec:knowledgegradient}
The knowledge gradient is the marginal value of an experiment, measured in terms of its ability to make better decisions and produce a better design.  For example, imagine we have a function (which could be a lookup table, linear model, or sampled nonlinear model) $F(y)$, and let $\Fbar^n(y)$ be our estimate of the function after $n$ experiments.  If we stop now, we would choose as our design the value of $y^n=y$ that maximizes $\Fbar^n(y)$.  We could write that the current ``value'' of our problem (that is, the performance we obtain if we stop now) is given by
\bn
V^n(S^n) = \max_y \Fbar^n(y).  \label{eq:vn}
\en
where $S^n$ is our current state of knowledge (which means both the point and distributional estimates of the true function $F(y)$).

Now assume we are thinking about running an experiment with controls $x$, and then obtaining an updated estimate $F^{n+1}(y|x)$.  Since we have not yet run this experiment, we do not yet know the outcome of the experiment, so we have to view this as a random variable.  Imagine that we run this $i=1, \ldots, 100$ times, and let $W^{n+1}_i$ be the experimental outcome from the $ith$ experiment.  We then use this information to update our function (using the methods we described in section \ref{sec:beliefmodels}), giving us 100 updated estimates of our belief model $F^{n+1}_i(y|x)$.  Finally, let $y^{n+1}_i$ be the optimal design based on the outcome of the $ith$ version of our experiment, where $y^{n+1}_i = \argmax_y F^{n+1}_i(y|x)$.  The value of running an experiment with controls $x$, when averaged across all 100 repetitions, is given by
\bn
Q^{n+1}(S^n,x) = \frac{1}{100}\sum_{i=1}^{100} F^{n+1}_i(y^{n+1}_i|x).  \label{eq:vnplus1}
\en

We can compute the marginal value of running experiment with controls $x$ by looking at how much better $Q^{n+1}(S^n,x)$ is than $V^n(S^n)$ by subtracting \eqref{eq:vn} from \eqref{eq:vnplus1}, giving us
\bn
\nu^{KG,n}(x) = Q^{n+1}(S^n,x) - V^n(S^n). \label{eq:knowledgegradient}
\en
The quantity $\nu^{KG,n}(x)$ is the marginal value (on average) of running an experiment with controls $x$.  This is known in the literature as the {\it knowledge gradient}, which gives us the expected value of running experiment $x$ in terms of how well it improves our design of our material, drug or device.  The idea is to compute $\nu^{KG,n}(x)$ for each possible experiment $x$ (and this may be a lot), and then choose the experiment that best improves the value of information.

\subsection{Knowledge gradient for lookup table beliefs}
The knowledge gradient for independent beliefs can be easily computed in a spreadsheet; one can be downloaded from:\\
\indent \url{http://optimallearning.princeton.edu/software/KnowledgeGradient.xlsx}.  \\
We present without derivation the set of formulas for calculating the knowledge gradient for an experiment $x$, where each possible experiment $x$ has a belief about the true mean $\mu_x$ which is normally distributed with mean $\theta^n_x$ and precision (the inverse variance) of $\beta^n_x$.  We then use this simple model to derive some insights into the properties of the knowledge gradient.

We start by computing $\sigmatilde^{2,n}_x$ which is the variance in $\theta^{n+1}_x$ given what we know at time $n$:
\bn
\sigmatilde^{2,n}_x &=& (\beta^n_x)^{-1} - (\beta^n_x + \beta^W)^{-1}. \label{eq:kgsigmatilde}
\en
We then compute the normalized distance between $\theta^n_x$ (that is, what we think is the performance of design $n$) and the next best alternative (we then turn this into a negative number - don't ask):
\bn
\zeta^n_x &=& -\left|\frac{\theta^n_x - \max_{x'\ne x} \theta^n_{x'}}{\sigmatilde^n_x}\right|.  \label{eq:KGzeta}
\en
Finally, we compute the function
\bn
f(\zeta) = \zeta \Phi(\zeta) + \phi(\zeta),  \label{eq:fzeta}
\en
where $\Phi(\zeta)$ and $\phi(\zeta)$ are, respectively, the cumulative standard normal distribution and the standard normal density.  That is,
\bns
\phi(\zeta) = \frac{1}{\sqrt{2\pi}} e^{-\frac{\zeta^2}{2}},
\ens
and
\bns
\Phi(\zeta) = \int_{-\infty}^\zeta \phi(x) dx.
\ens
There are functions for computing $\Phi(\zeta)$ in most programming languages.  For example, in Excel this is \text{NORMSDIST($\zeta$)}.  Finally, the knowledge gradient is given by
\bn
\nu^{KG,n}_x = \sigmatilde^n_x f(\zeta^n_x). \label{eq:KG}
\en

\begin{figure}[tb]
  \begin{center}
  \begin{tabular}{ccc}
  \includegraphics[width=2.0in]{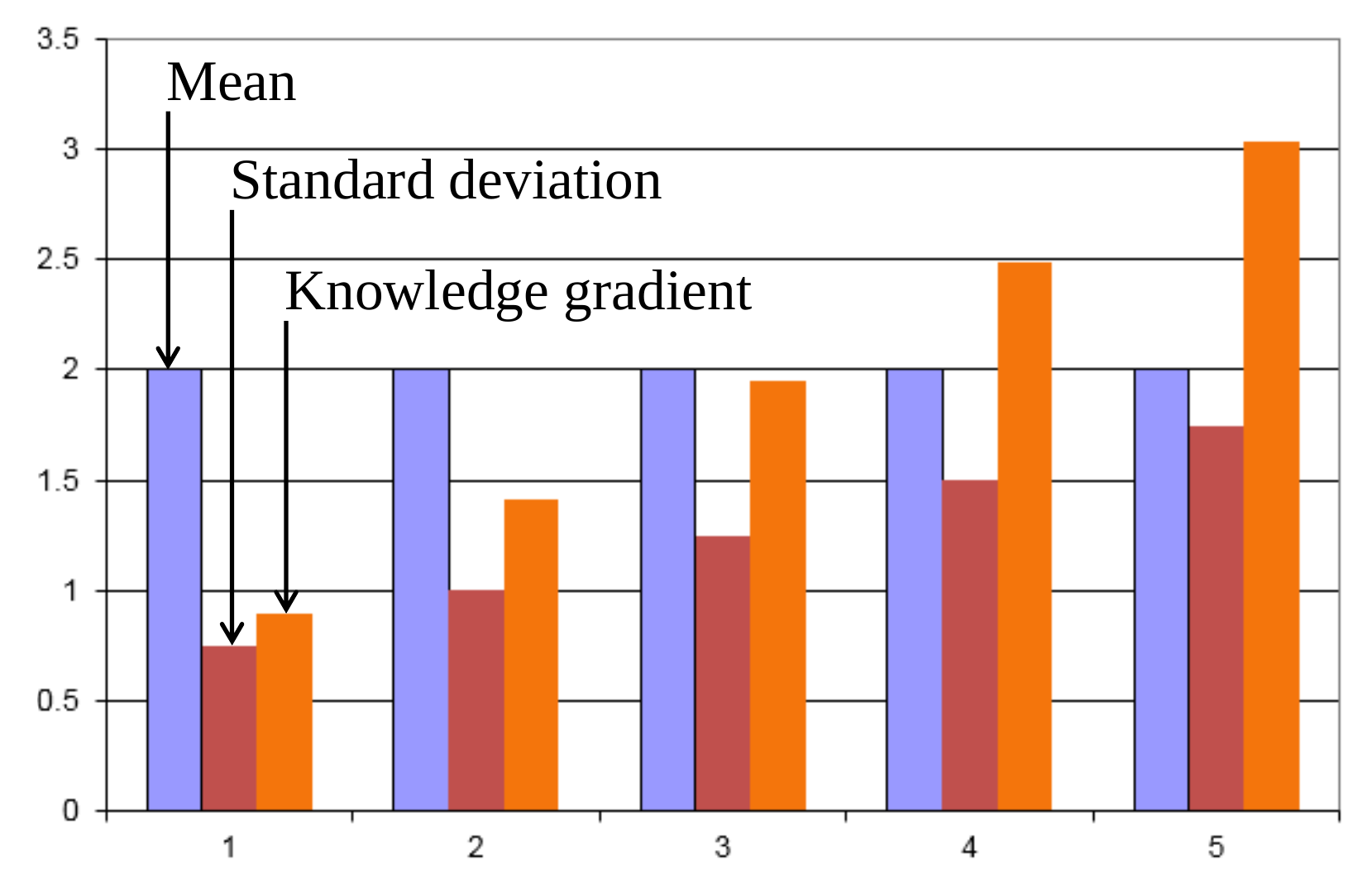} & \includegraphics[width=2.0in]{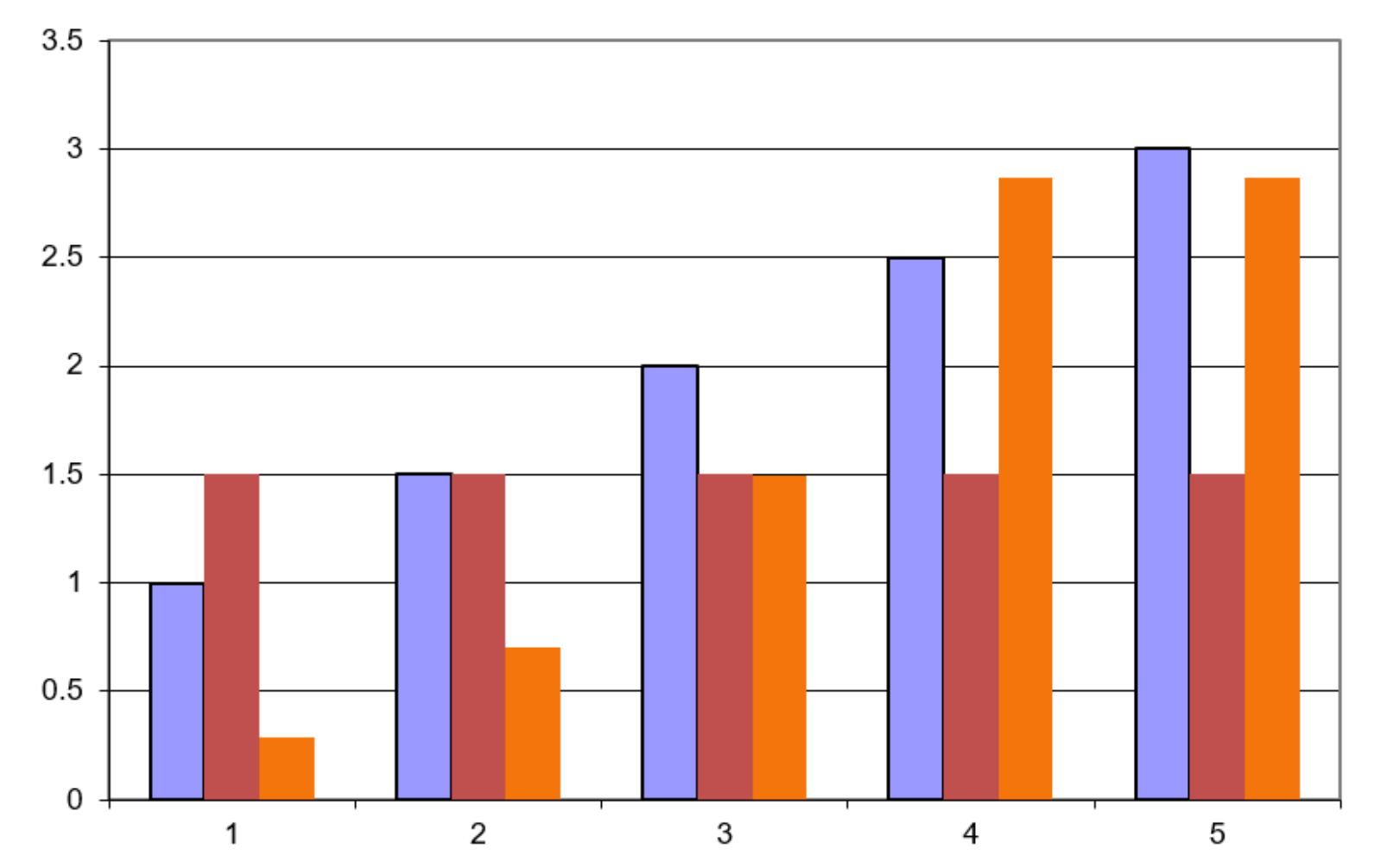} & \includegraphics[width=2.0in]{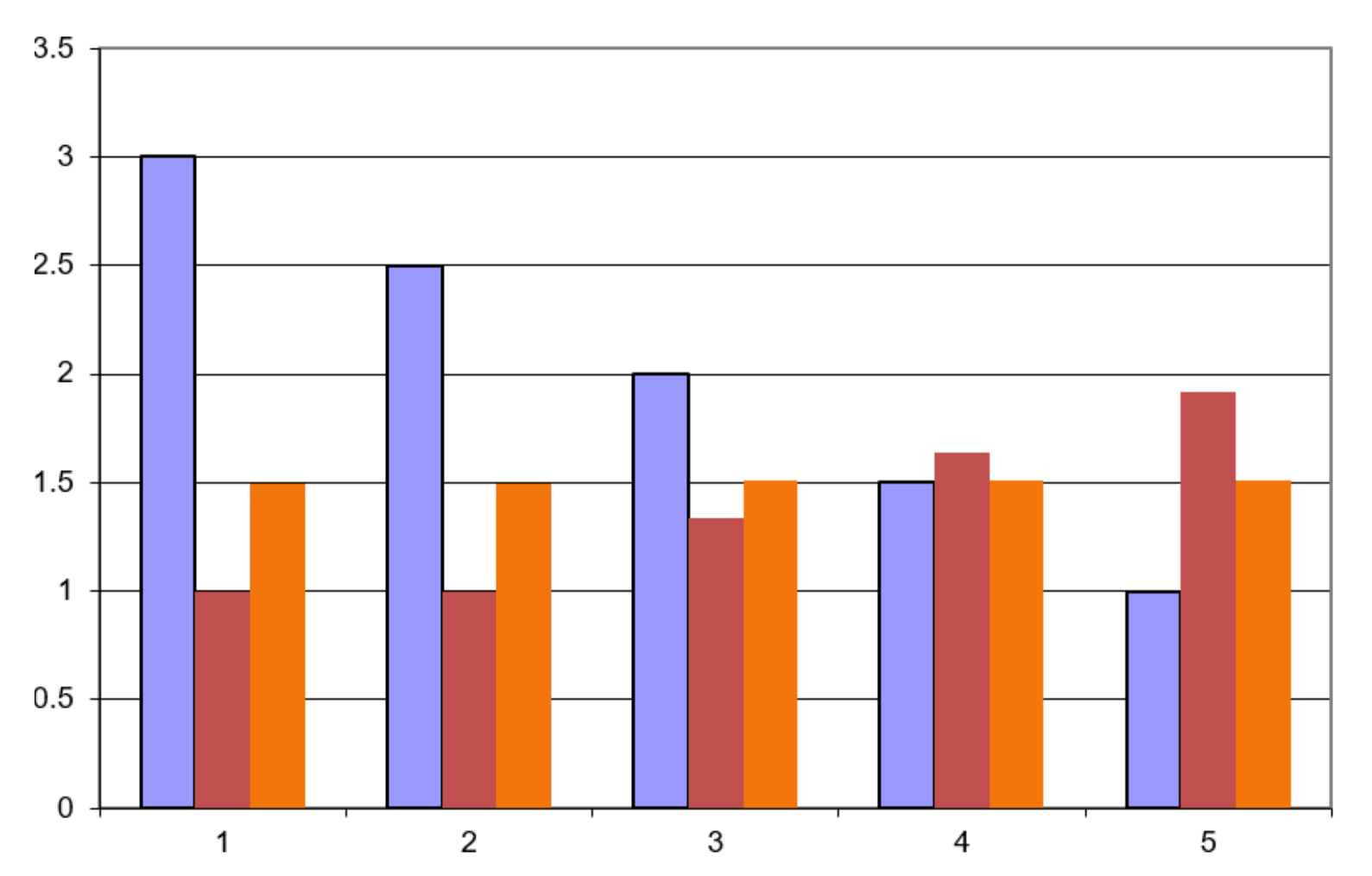}\\
                            (a)          &                (b)           &    (c)
  \end{tabular}
  \end{center}
  \caption{The knowledge gradient for lookup table with independent beliefs with equal means (a), equal variances (b), and adjusting means and variances so that the KG is equal (c).}\label{fig:KGequal}
\end{figure}

While lookup table models with independent beliefs are unlikely to be of much practical value for realistic problems that arise in laboratory settings, we can derive some useful insights from these calculations.  Figure \ref{fig:KGequal} shows the means, standard deviations and the knowledge gradient (scaled to make the graphs more readable), for three sets of problems with five alternatives.  We observe the following:
\begin{itemize}
\item Figure \ref{fig:KGequal}(a) depicts a problem where the means are all equal, with increasing standard deviations, which shows that as the uncertainty increases, so does the knowledge gradient (in other words, all else held equal, it is better to do experiments where there is more uncertainty).
\item Figure \ref{fig:KGequal}(b) depicts a problem where the standard deviations are all equal, showing that when the uncertainty is the same, KG increases with the mean (bigger is better).
\item Figure \ref{fig:KGequal}(c) adjusts the mean and variance so that the knowledge gradient is equal, showing the tradeoff between how good an experiment looks, and the uncertainty.
\end{itemize}

This tradeoff between the expected performance of an experiment, and the uncertainty, has turned out to be a critical feature of the best policies.  For example, we have found that a properly tuned interval estimation policy (equation \eqref{eq:intervalestimation}) can work extremely well (the key qualifier is ``properly tuned'').  Note that interval estimation has an explicit tradeoff between how promising a design looks (given by $\theta^n_x$), and how uncertain we are (given by $\sigma^n_x$).  This is an insight that all scientists should keep in mind when running experiments - uncertainty is good (but within reason).

\subsection{Properties of the value of information}
The value of information is a powerful idea in the setting of expensive laboratory experiments.  It makes intuitive sense to run the experiment where you learn the most (although in practice scientists often tend to run the experiment that they feel is most likely to be successful). However, the value of information has to be used with some care.

\begin{figure}[tb]
  \begin{center}
  \begin{tabular}{cc}
  \includegraphics[width=3.0in]{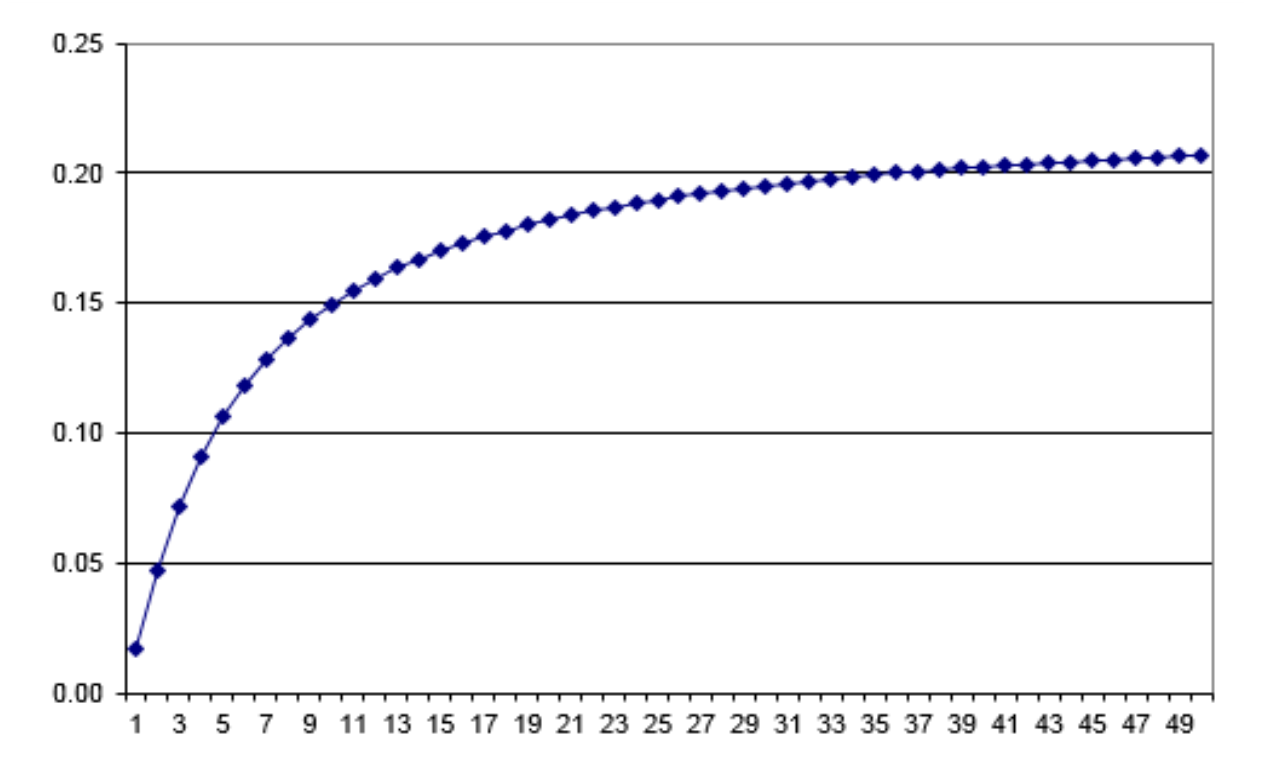} & \includegraphics[width=3.0in]{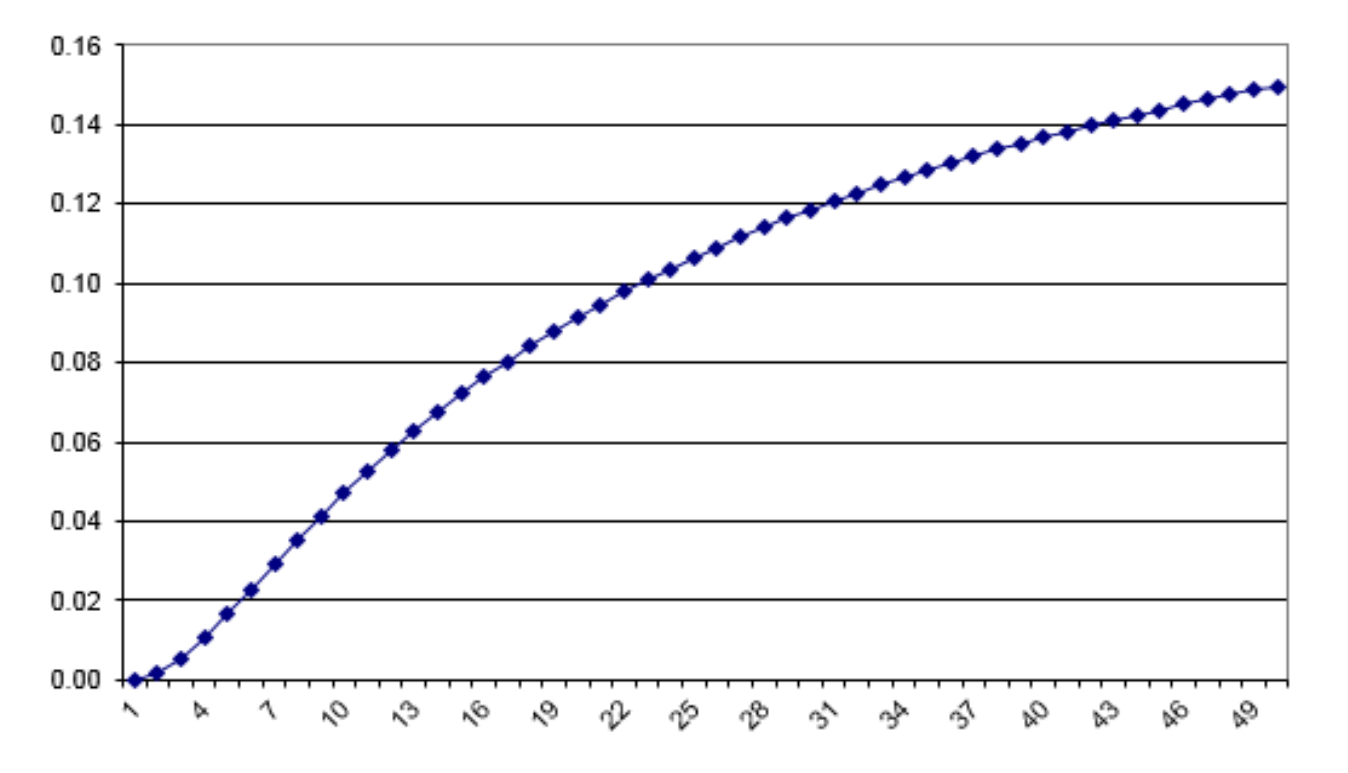} \\
                            (a)                      &                           (b)
  \end{tabular}
  \end{center}
  \caption{Illustration of the value of information is concave (a) versus an S-curve (b) with respect to the number of experiments.}\label{fig:valueofinformation}
\end{figure}

Imagine running an experiment $x$, and then repeating this same experiment over and over.  It is reasonable to expect that the value of information from each additional experiment would decline as experiments are repeated. It turns out that this is often true, but only when experiments are not too noisy.  Figure \ref{fig:valueofinformation}(a) illustrates the value of information from $k$ experiments as $k$ is increased from 1 to 100.  This figure illustrates a situation where the value of information is concave in the number of times an experiment is repeated.

Figure \ref{fig:valueofinformation}(b), however, illustrates a case where this is not true.  This situation arises when experiments are noisy, which means that we simply do not learn much from a single experiment.  One situation where this is almost guaranteed to happen is when the outcome of an experiment is of the success/failure type.  Making judgments from a single experiment with a success/failure outcome is like deciding if a baseball player is a good hitter after one at bat.

It is fairly easy to identify if the value of information is concave or not.  All you have to do is to calculate the knowledge gradient, replacing the precision of an experiment $\beta^W$ with a repetition factor, such as $k \beta^W$ (this would be done in equations \eqref{eq:bayesmeanx} and \eqref{eq:bayesprecisionx}).  If the knowledge gradient increases when $k$ is increased from 1 to 2, then your value of information exhibits the S-curve behavior shown in \ref{fig:valueofinformation}(b).  In this case, using the knowledge gradient with a repetition factor $k=1$ provides very little guidance.

This situation is easy to fix.  Let $\nu^{KG,n}_x(k)$ be the knowledge gradient with a repetition factor of $k$. Now find $k$ that maximizes $\nu^{KG,n}_x(k)/k$ (think of this as maximizing the average value of information). Do this for each possible experiment $x$, and then do the experiment with the highest {\it average} value of information (this idea was suggested as the KG(*) policy in \cite{Frazier2010}).  Note that we do not actually repeat the experiment $k$ times; we just use this average value of an information to guide the {\it next} experiment we should perform.

\subsection{The knowledge gradient for online learning}
The knowledge gradient is the value of information if we perform one more experiment, and live with those results.  We can also think of this as the improvement in our final design after we have exhausted our budget of $N$ experiments (we referred to this above as the ``final reward'' objective).  In laboratory sciences, it seems intuitive to focus on the final reward, because this is the design we take to the field.  This means we do not mind experiments that work poorly, as long as we learn as much as possible, leading to a good final solution.

As we have worked in this problem domain, we have come to appreciate the importance of doing well even while we are in the lab.  The reason is that offline learning, especially with a parametric belief model (any parametric model) often leads to doing extreme experiments.  For example, if we are trying to learn which of the set of uncertain lines in figure \ref{fig:correlatedlines} is correct, the best experiments to run are at the extremes. In practice, not only might these experiments be hard to run, the more serious problem is that our parametric model is typically just an approximation which is likely to fit the best in the region of the optimum.  For example, molecules ``denature'' at lower and higher temperatures which mean that idealized models of interactions are less accurate at extremes.  Also, high temperatures can bring out unmodeled kinetic processes such as gas-phase pyrolysis of carbon.  As a result, we do not learn as much from extreme experiments.

It is very easy to convert the offline knowledge gradient to an online version.  If $\nu^{KG,n}_x$ is the expected value of the information gained from running an experiment with controls $x$ (that is, this is the improvement in the final objective after $N$ experiments), then we can construct an online version using
\bn
\nu^{OLKG,n}_x = \theta^n_x + (N-n) \nu^{KG,n}_x.  \label{eq:onlinekg}
\en
Here, we are adding the performance we think we will achieve in the next experiment, given by $\theta^n_x$, plus the marginal increase in the improvement in future decisions from the next experiment, times the number of remaining experiments, given by $N-n$.

The online knowledge gradient has some nice properties.  As with the interval estimation policy, it has a nice balance between doing the best we can now (exploitation), because of the term $\theta^n_x$, and learning for the future (exploration), through the term $(N-n) \nu^{KG,n}_x$.  Further, the exploration term starts off larger, and steadily shrinks as we get close to the end of our budget (we are unaware of any other formal learning policy that does this).  Finally, there are no tunable parameters; instead, the knowledge gradient takes advantage of the expertise of the scientist through the prior.

\subsection{The advantages of the knowledge gradient}
We have considerable experience comparing the performance of a wide range of learning policies over a decade-long research effort (as of this writing).  In particular we have worked in the context of laboratory sciences, which offers unique characteristics compared to other learning settings. With this backdrop, we have found the most attractive features of the knowledge gradient to include:
\begin{itemize}
   \item The property that the knowledge gradient maximizes the marginal value of information when the possibility of an S-curve behavior is properly handled.  This makes the knowledge gradient better suited for small experimental budgets.
   \item The ability to handle a wide range of belief models, including:
   \begin{itemize}
      \item Lookup table with correlated beliefs.
      \item Linear parametric models, including high-dimensional sparse additive belief models (linear models with hundreds of parameters, where most are zero).
      \item Nonlinear parametric models where we have experience with both sampled belief models (\cite{Chen2015}, \cite{He2016}) and the Laplace approximation (\cite{Wang2016},\cite{Wang2016a}) for the logistic regression belief model.
      \item A variety of nonparametric belief models, including hierarchical beliefs (ideal for problems with a large number of discrete choices), kernel regression (which smooths between nearby points), and locally linear approximations.
   \end{itemize}
   \item The value of information can be provided as an input to the decision-making process by scientists (since we can calculate the KG score for all alternatives), allowing them to make other tradeoffs that are not fully captured by our models.
   \item The ability to handle both offline (final performance) and online (cumulative performance) objectives.  This capability helps to overcome the current tendency of offline policies to test the experimental boundaries when using parametric belief models.
   \item No tunable parameters - Policies that fall under the heading of policy search require parameter tuning, which can only be done in a simulator.  The lack of tunable parameters is ideal for experimental science with little to no data, but where domain knowledge can be used to build a prior.
\end{itemize}

\section{Implementing recommendations}
\label{sec:implementingrecommendations}
We have now reviewed a wide range of strategies for guiding the experimentation process.  Often the real challenge is getting scientists to actually implement these recommendations.  Below we discuss the implementation process from the perspective of humans guiding the process versus machines in the form of robotic scientists.

\subsection{Working with people}
\label{sec:workingwithpeople}
\begin{figure}[tb]
  \center{\includegraphics[width=5.0in]{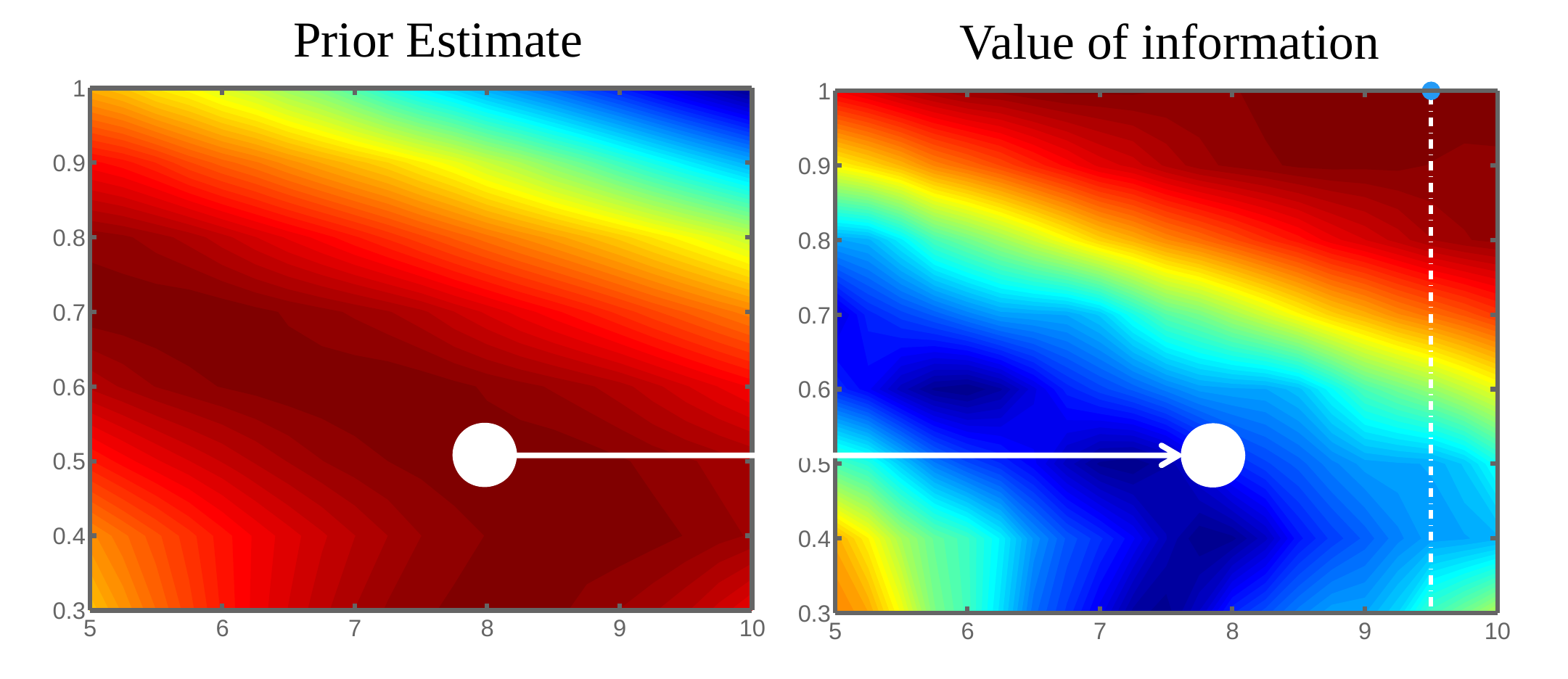}}\\
  \caption{On the left is the expected performance if we run an experiment at the indicated location. On the right is the value of information gained, which is important if the experiment fails.}\label{fig:exploitationvsexploration}
\end{figure}

The value of information has proven to be particularly useful when we have to interact with a scientist who makes the final decision.  The left side of figure \ref{fig:exploitationvsexploration} shows a heat map giving the regions where we anticipate the best outcomes (for example, we will be able to achieve the highest conductivity or material strength).  It makes sense to run an experiment which seems as if it is likely to be successful.  However, not all experiments are successes, so we need to think about how much we would learn with this experiment in case it fails (which is most of the time).  The heat map to the right shows the value of information from each experiment, which indicates we will learn very little.

Since many experiments are not successful, it might make sense to run the experiment where we can learn the most, which would be one of the red regions in the heat map on the right.  However, this may not be best either. In this heat map (obtained when using a parametric model), we see familiar behavior where the highest value of information is on the borders of the figure.  This might correspond to experiments being run at very high temperatures or concentrations.  These may be hard to run, or we may have doubts about the accuracy of our parametric model in these regions.  This is where scientific judgment can play a role.

\begin{figure}[tb]
  \center{\includegraphics[width=5.0in]{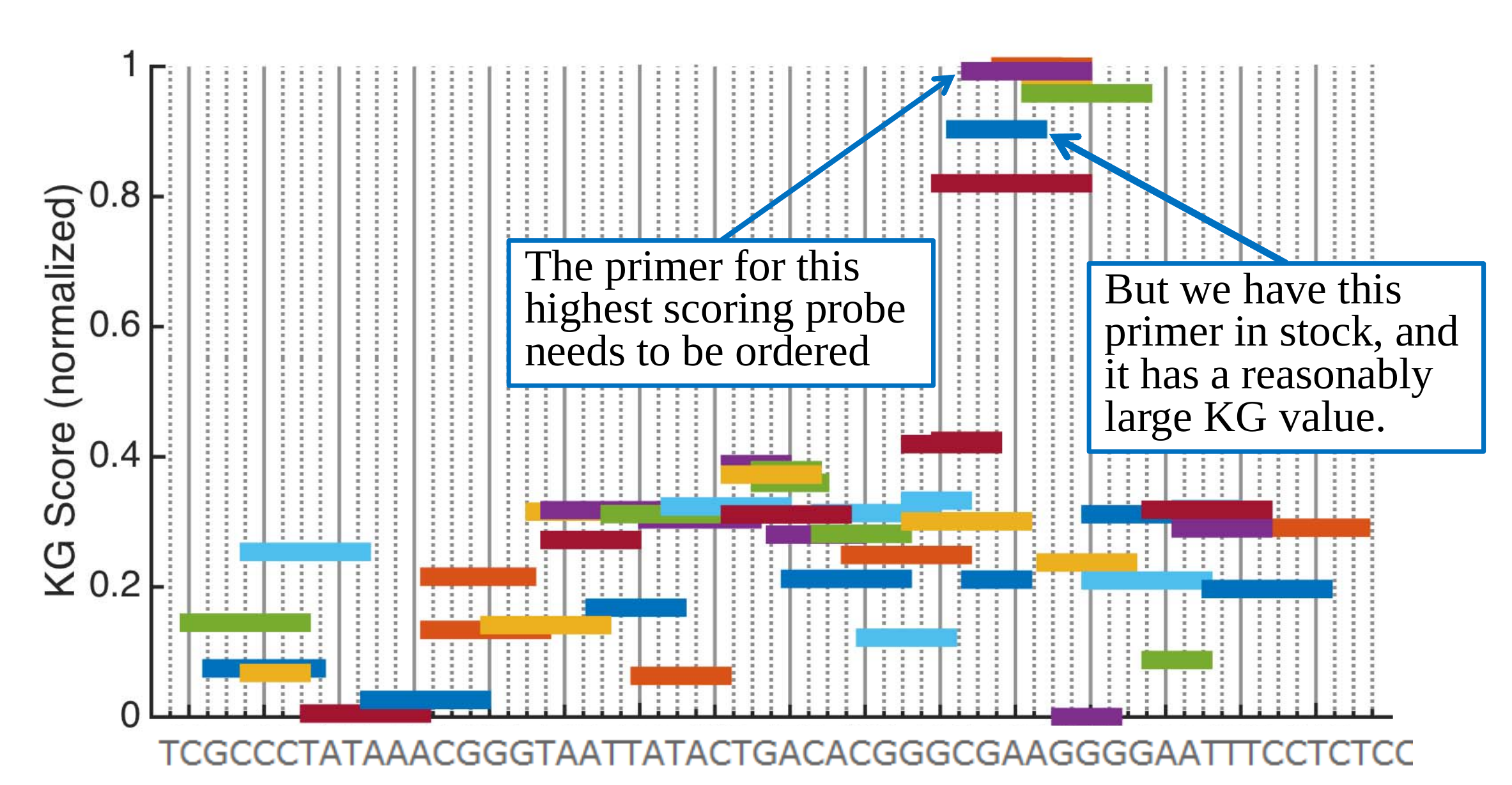}}
  \caption{The value of probes designed for different regions of an RNA molecule, showing the value of information of each.}\label{fig:probe}
\end{figure}

Another example is shown in figure \ref{fig:probe}, which depicts the value of information for probes designed to attach to different segments of an RNA molecule.  There is a group of probes with a much higher value of information than the others.  However, of these, some are easier for a scientist to manufacture because of the materials on hand in the lab.  This information is not known to the knowledge gradient calculation, but the graphic makes it possible for scientists to balance the value of information (and likely performance) against other issues.

\subsection{Working with machines}
\label{sec:workingwithmachines}
Robotic scientists offer a unique opportunity for analytics-based ``high-throughput'' experimentation, since they are completely dependent on analytics for guiding the experimental process.  Some issues that have to be managed with robots include
\begin{itemize}
\item Robotic scientists work far more quickly than humans, as long as experiments are defined within a specific range.  This capability makes brute force solutions easier, and reduces the value of developing accurate priors (although these would dramatically accelerate a robotic scientist).
\item Robots make it easier to do certain types of experiments in high volume (such as varying well-defined continuous parameters).  Other changes, such as moving to a new catalyst, will seem slow by comparison.
\item Changes in the experimental process itself are slower because it may require changes to the robot itself.
\item We can no longer depend on the expertise of a scientist to compensate for limitations in the mathematical policy.
\end{itemize}
Robotic scientists represent a new frontier for laboratory research, with open questions in terms of how robotics are best blended with human decisions.  We anticipate that this will offer new challenges for optimal learning.




\section{Assessing risk}
\label{sec:assessingrisk}
Whether using the knowledge gradient or any other form of learning policy, it is now possible to run simulations by letting the analytic policy make decisions about which experiment to run next.  This provides us with a powerful methodology for assessing the risk associated with an experimental series.

\begin{figure}[tb]
  \center{\includegraphics[width=4.0in]{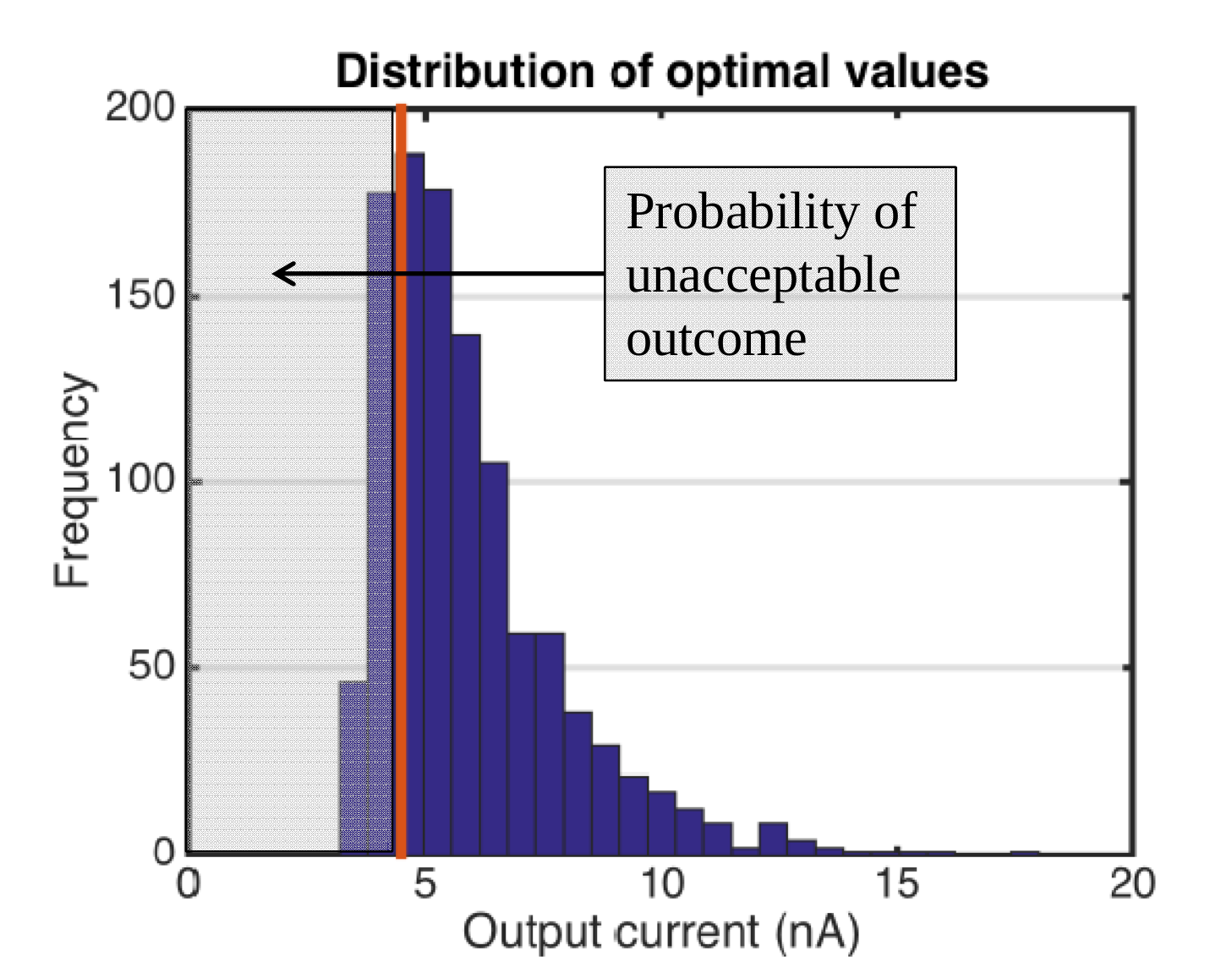}}
  \caption{Distribution of the results of a sequence of experiments, simulated 100 times, showing the number of times the experiments did not achieve the desired target.}\label{fig:risk}
\end{figure}

In a real lab, we can only execute a series of experiments once.  In the computer, we can simulate a series of experiments 1,000 times, and then see how often the experiments result in a design that meets or exceeds some threshold.  Using an experimental policy $\pi$ (which can be any of the policies that we have discussed above) given a prior belief $S^0$, let $F^\pi_i(S^0)$ be the performance of the $ith$ simulation.  This performance might be what we believe we have achieved in terms of the strength of a material, conductivity of a surface, or the number of cancer cells killed. If we repeat this simulation 1,000 times, we can create a distribution such as that shown in figure \ref{fig:risk}.

A frequency diagram such as that given in \ref{fig:risk} can be used to assess how the experimental process is designed affects risk.  These include:
\begin{itemize}
\item Budget (time and cost) - More experiments increase the likelihood that you will achieve a particular objective.
\item The prior - More time spent on getting a better prior (for example, spending several months on a thorough literature review or running numerical simulations to improve our understanding of a molecule) will produce better results, but the time and money spent developing the prior may not be worth it.
\item Testing equipment - Better equipment may reduce experimental variation.
\item The experimental process - For example, we might consider the investment in using a robotic scientist, or experiment with different sequencing of steps.
\item The choice of experimental policy.
\end{itemize}
Simulations to assess the risk of a sequence of experiments requires using judgment to estimate inputs such as the quality of the prior, the variability in the experimental process, and the time and cost of different experimental steps.  While errors in these estimates are to be expected, we think this type of risk analysis can be undertaken relatively easily, and should provide an indication of the value of larger budgets, better initial estimates (the prior) and better equipment.  Further, we feel that the exercise of understanding the different types of uncertainty will help to inform experimental decisions.

\section{Concluding remarks}
\label{sec:concludingremarks}
The intent of this tutorial is to provide laboratory scientists and research managers with an introduction to the science of designing and executing experiments.  Some of the most important points include:
\begin{itemize}
\item The importance of modeling uncertainty in your understanding of the physical process, which translates to uncertainty in the performance $F(x)$ when you run an experiment with controls $x$.
\item Recognizing that the point of an experiment is to reduce uncertainty in a way that leads to better decisions about the final design.  While it is tempting to shoot for the big win, most experiments fail, and it is important to think about what you will learn from each experiment, and how this will help decisions in the future.
\end{itemize}

At the same point, it is important to recognize what we have not considered.  It has been our experience that the biggest challenge facing scientists are the decisions we are not even considering.  We talk about running an experiment with controls $x$, but what about the choices we are not even considering?  It should not be surprising that scientists naturally gravitate toward doing the experiments that are easiest.  Once scientist described the possibility of designing and building an entirely new machine, which would take a year.  Less dramatic is the choice of stepping back and letting a robotic scientist run 100 experiments over several hours, versus taking a day to switch to a new catalyst.

%
%
%


\section*{Acknowledgements}
This research was supported in part by AFOSR grant contract FA9550-12-1-0200 for Natural Materials, Systems and Extremophiles and the program in Optimization and Discrete Mathematics.



\singlespace
\addcontentsline{toc}{section}{References}



\end{document}